\pdfoutput=1

\documentclass{article}
\usepackage{times}

\usepackage{url}
\usepackage[utf8]{inputenc} 
\usepackage{geometry} 
\usepackage{amsmath, amssymb, graphicx}
\usepackage{bbm}
\usepackage{algorithm,algorithmic}
\usepackage{hyperref}

\newcommand{\EE}{\mathbb{E}}

\newcommand{\PP}{\mathbb{P}}
\newcommand{\RR}{\mathbb{R}}
\newcommand{\indic}{\textbf{1}}

\newcommand{\bp}{\noindent{\textbf{Proof.}}\ }
\newcommand{\ep}{\hfill $\Box$}

\newcommand{\al}[1]{ \begin{align} #1  \end{align}}
\newcommand{\eq}[1]{ \begin{equation} #1  \end{equation}}
\newcommand{\als}[1]{ \begin{align*} #1  \end{align*}}
\newcommand{\eqs}[1]{ \begin{equation*} #1  \end{equation*}}

\newcommand{\sk}{\nonumber\\} 
\newcommand{\Lp}{\left(}
\newcommand{\Rp}{\right)}

\newcommand{\el}{\end{flushleft}}
\newcommand{\bl}{\begin{flushleft}}
\newcommand{\floor}[1]{  \lfloor #1 \rfloor} 
\newcommand{\ceil}[1]{  \lceil #1 \rceil} 

\newcommand{\ouralgo}{SP }
\newcommand{\ouralgom}{SP} 
 
\newcommand{\subalgom}{IT} 
\newcommand{\ouralgoprime}{SP' } 
\newcommand{\ouralgoprimem}{SP'} 
\newcommand{\subalgoprimem}{IT'}

\newcommand{\KL}{ \text{KL }} 
\newcommand{\KLtwo}{ \text{KL}_2 }

\newcommand{\separator}{
  \begin{center}
    \rule{\columnwidth}{0.3mm}
  \end{center}
}

\newtheorem{proposition}{Proposition}
\newtheorem{theorem}{Theorem}[section]
\newtheorem{lemma}[theorem]{Lemma}
\newtheorem{corollary}[theorem]{Corollary}
\newtheorem{remark}{Remark}

\newtheorem{definition}[theorem]{Definition}	

\DeclareGraphicsExtensions{.pdf,.eps}

\graphicspath{{.//}}

\title{Unimodal Bandits without Smoothness}

\author{Richard Combes \thanks{Supelec, France, mail: richard.combes@supelec.fr} \hspace{0.2cm} and Alexandre Prouti\`ere \thanks{KTH, Sweden, mail: alepro@kth.se}}

\begin{document}

\maketitle

\begin{abstract}
We consider stochastic bandit problems with a continuous set of arms and where the expected reward is a continuous and unimodal function of the arm. No further assumption is made regarding the smoothness and the structure of the expected reward function. For these problems, we propose the Stochastic Pentachotomy (SP) algorithm, and derive finite-time upper bounds on its regret and optimization error. In particular, we show that, for any expected reward function $\mu$ that behaves as $\mu(x)=\mu(x^\star)-C|x-x^\star|^\xi$ locally around its maximizer $x^\star$ for some $\xi, C>0$, the SP algorithm is order-optimal. Namely its regret and optimization error scale as $O(\sqrt{T\log(T)})$ and $O(\sqrt{\log(T)/T})$, respectively, when the time horizon $T$ grows large. These scalings are achieved without the knowledge of $\xi$ and $C$. Our algorithm is based on asymptotically optimal sequential statistical tests used to successively trim an interval that contains the best arm with high probability. To our knowledge, the SP algorithm constitutes the first sequential arm selection rule that achieves a regret and optimization error scaling as $O(\sqrt{T})$ and $O(1/\sqrt{T})$, respectively, up to a logarithmic factor for non-smooth expected reward functions, as well as for smooth functions with unknown smoothness.  
\end{abstract}

\section{Introduction} \label{sec:intro}

This paper considers the problem of stochastic unimodal optimization with bandit feedback which is a generalization of the classical multi-armed bandit problem solved by Lai and Robbins \cite{lai1985}. The problem is defined by a continuous and unimodal expected reward function $\mu$ defined on the interval $[0,1]$. For this problem, we consider  algorithms that repeatedly select an arm $x\in [0,1]$, and get a noisy reward of mean $\mu(x)$. The performance of an algorithm is characterized by its regret and its optimization error up to time horizon $T$ (the number of observed noisy rewards). The regret is the difference between the average cumulative reward one would obtain if the function $\mu$ was known, i.e., $T\sup_{x\in [0,1]}\mu(x)$, and the actual average cumulative reward achieved under the algorithm. The optimization error is the difference between $\sup_{x\in [0,1]}\mu(x)$ and the expected reward of the arm selected at time $T$. Known lower bounds for the regret and optimization error scale as $\Omega(\sqrt{T})$ (for linear reward functions) and $\Omega(1/\sqrt{T})$ (for quadratic reward functions), respectively. Our objective is to devise an algorithm whose regret and optimization error scale as $O(\sqrt{T})$ and $O(1/\sqrt{T})$ up to a logarithmic factor for a large class of unimodal and continuous reward functions. Such an algorithm would hence be order-optimal. Importantly we merely make any assumption on the smoothness of the reward function -- the latter can even be non-differentiable. This contrasts with all existing work investigating similar continuum-armed bandit problems, and where strong assumptions are made on the structure and smoothness of the reward function. These structure and smoothness are known to the decision maker, and are explicitly used in the design of efficient algorithms.

We propose Stochastic Pentachotomy (SP), an algorithm for which we derive finite-time upper bounds on regret and optimization error. In particular, we show that its regret and optimization error scale as $O(\sqrt{T\log(T)})$ and $O(\sqrt{\log(T)/T})$ for any unimodal and continuous reward function $\mu$ that behaves as $\mu(x)=\mu(x^\star)-C|x-x^\star|^\xi$ locally around its maximizer $x^\star$ for some $\xi, C>0$. These scalings are achieved without the knowledge of $\xi$ or $C$, i.e., without the knowledge of the smoothness of $\mu$. The SP algorithm consists in successively narrowing an interval in $[0,1]$ while ensuring that the arm with the highest mean reward remains in this interval with high probability. The narrowing subroutine is a sequential test that takes as input an interval and samples a few arms in the interior of this interval until it gathers enough information to actually reduce the interval. We investigate a general class of such sequential tests. In particular, we provide a (finite time) lower bound of their expected sampling complexity given some guaranteed minimax risk, and design a sequential test that matches this lower bound. This optimal test is used in the SP algorithm. Interestingly, we show that to be efficient, a sequential test needs to sample at least three arms in the interior of the interval to reduce. This implies that a stochastic version of the celebrated Golden section search algorithm cannot achieve a reasonably low regret or optimization error over a large class of reward functions. Indeed such an algorithm would sample only two arms in the interval to reduce. We illustrate the performance of our algorithms using numerical experiments and compare its regret to that of existing algorithms that leverage the smoothness and structure of the reward function.

To our knowledge, SP is the first algorithm for continuous unimodal bandit problems that is order-optimal for a large class of expected reward functions: Its regret and optimization error scale as $O(\sqrt{T\log(T)})$ and $O(\sqrt{\log(T)/T})$ for non-smooth reward functions, as well as for smooth functions with unknown smoothness. 

\textbf{ Related work.} Stochastic bandit problems with a continuous set of arms have recently received a lot of attention. Various kinds of structured reward functions have been explored, i.e., linear \cite{dani08}, Lipschitz \cite{agrawal95}, \cite{kleinberg2008}, \cite{bubeck08}, and convex \cite{agarwal2013}, \cite{shamir2013}. In these papers, the knowledge of the structure greatly helps the design of efficient algorithms (e.g. for Lipschitz bandits, except in \cite{bubeck2011alt}, the Lipschitz constant is assumed to be known). More importantly, the smoothness or regularity of the reward function near its maximizer is also assumed to be known and leveraged in the algorithms. Indeed, most existing algorithms use a discretization of the set of arms that depends on this smoothness, and this is crucial to guarantee a regret scaling as $O(\sqrt{T})$. As discussed in \cite{bubeck08}, \cite{auer2007}, without the knowledge of the smoothness, these algorithms would yield a much higher regret (e.g. scaling as $O(T^{2/3})$ for the algorithm proposed in \cite{auer2007}).

Unimodal bandits with a continuous set of arms have been addressed in \cite{cope09}, \cite{yu2011}. In \cite{cope09}, the author shows that Kiefer-Wolfowitz (KW) stochastic approximation algorithm achieves a regret of the order of $O(\sqrt{T})$ under some strong regularity assumptions on the reward function (strong convexity). LSE, the algorithm proposed in \cite{yu2011}, has a regret that scales as $O(\sqrt{T}\log(T))$, but requires the knowledge of the smoothness of the reward function. LSE is a stochastic version of the Golden section search algorithm, and iteratively eliminates subsets of arms based on PAC-bounds derived after appropriate sampling. By design, under LSE, the sequence of parameters used for the PAC bounds is pre-defined, and in particular does not depend of the observed rewards. As a consequence, LSE may explore too much sub-optimal parts of the set of arms. Our algorithm exploits more adaptive sequential statistical tests to remove subsets of arms, and yields a lower regret even without the knowledge of the smoothness of the reward function. A naive way to address continuous-armed bandit problems consists in discretizing the set of arms, and in applying efficient discrete bandit algorithms. This method was introduced in \cite{Kleinberg2004}, and revisited in \cite{combes2014} in the case of unimodal rewards. To get a regret scaling as $O(\sqrt{T}\log(T))$ using this method, the reward function needs to be smooth and the discretization should depend on the smoothness of the function near its maximizer. 

Our problem is related to stochastic derivative-free optimization problems where the goal is to get close to the maximizer of the reward function as quickly as possible, see e.g. \cite{spall2003}, \cite{jamieson2012}, and references therein. However, as explained in \cite{agarwal2013}, minimizing regret and optimization error constitute different objectives. Finally, it is worth mentioning papers investigating the design of sampling strategies to identify the best arm in multi-armed bandit problems, see e.g. \cite{mannor2004}, \cite{evendar2006}, \cite{audibert2010}, \cite{kalya2012}, \cite{jamieson2013}. These strategies apply to finite sets of arms, but resemble our sequential statistical tests used to reduce the interval containing the best arm. We believe that our analysis (e.g. we derive finite-time lower bounds for the expected sampling complexity of a set of tests), and our proof techniques are novel. 

\section{Problem Formulation and Notation}

We consider continuous bandit problems where the set of arms is the interval $[0,1]$, and where the expected reward $\mu$ is a continuous and unimodal function of the arm. More precisely, there exists $x^\star$ such that $x\mapsto \mu(x)$ is strictly increasing (resp. decreasing) in $[0,x^\star]$ (resp. in $[x^\star, 1]$). We denote by ${\cal U}$ the set of such functions. Define $\mu^\star = \mu(x^\star)$. 

Time proceeds in rounds indexed by $n=1,2,\ldots$. When arm $x$ is selected in round $n$, the observed reward $X_n(x)$ is a random variable whose expectation is $\mu(x)$ and whose distribution is $\nu(\mu(x))$, where $\nu$ refers to an exponential family of distributions with one parameter (e.g Bernoulli, exponential, Gaussian, ...). We assume that the rewards $(X_n(x),n\ge 1)$ are i.i.d., and are independent across arms. At each round, a decision rule or algorithm selects an arm depending on the arms chosen in earlier rounds, and the corresponding observed rewards. Let $x^\pi(n)$ denote the arm selected in round $n$ under the algorithm $\pi$. The set $\Pi$ of all possible algorithms consists of sequential decision rules $\pi$ such that for any $n\ge 2$, $x^\pi(n)$ is ${\cal F}_{n-1}^\pi$-measurable where ${\cal F}_n^\pi$ is the $\sigma$-algebra generated by $(x^\pi(s), X_s(x^\pi(s)), s=1,\ldots, n)$. The performance of an algorithm $\pi\in \Pi$ with time horizon $T$ is characterized by its regret $R^\pi(T)$ and optimization error $E^\pi(T)$ defined as $R^\pi(T) = T \mu^\star - \sum_{n=1}^T \EE[\mu( x^\pi(n) )]$ and $E^\pi(T) = \mu^\star - \EE[\mu( x^\pi(T))]$. Our objective is to devise an algorithm minimizing these performance metrics. Importantly, the only information available to the decision maker about the reward function $\mu$ is that $\mu\in {\cal U}$. In particular, the smoothness of $\mu$ around $x^\star$ remains unknown -- actually $\mu$ could well not be differentiable, e.g. $\mu(x)=\mu^\star-|x-x^\star |^\xi$ for $\xi \in (0,1)$.

\textbf{Notation.} In what follows, for any $\alpha, \beta$, we denote by $\KL(\alpha, \beta)$ the Kullback-Leibler divergence between distributions $\nu(\alpha)$ and $\nu(\beta)$. When $\alpha, \beta \in [0,1]$, and when $\nu(\cdot)$ is the family of Bernoulli distributions, this KL divergence is denoted by $\KLtwo(\alpha, \beta)=\KL(\alpha, \beta)=\alpha\log( \frac{\alpha}{\beta})+(1-\alpha)\log( \frac{1-\alpha}{1-\beta})$.

\section{Stochastic Polychotomy Algorithms}\label{sec:algo}

\begin{figure}
\begin{center}
	\includegraphics[width=0.8\columnwidth]{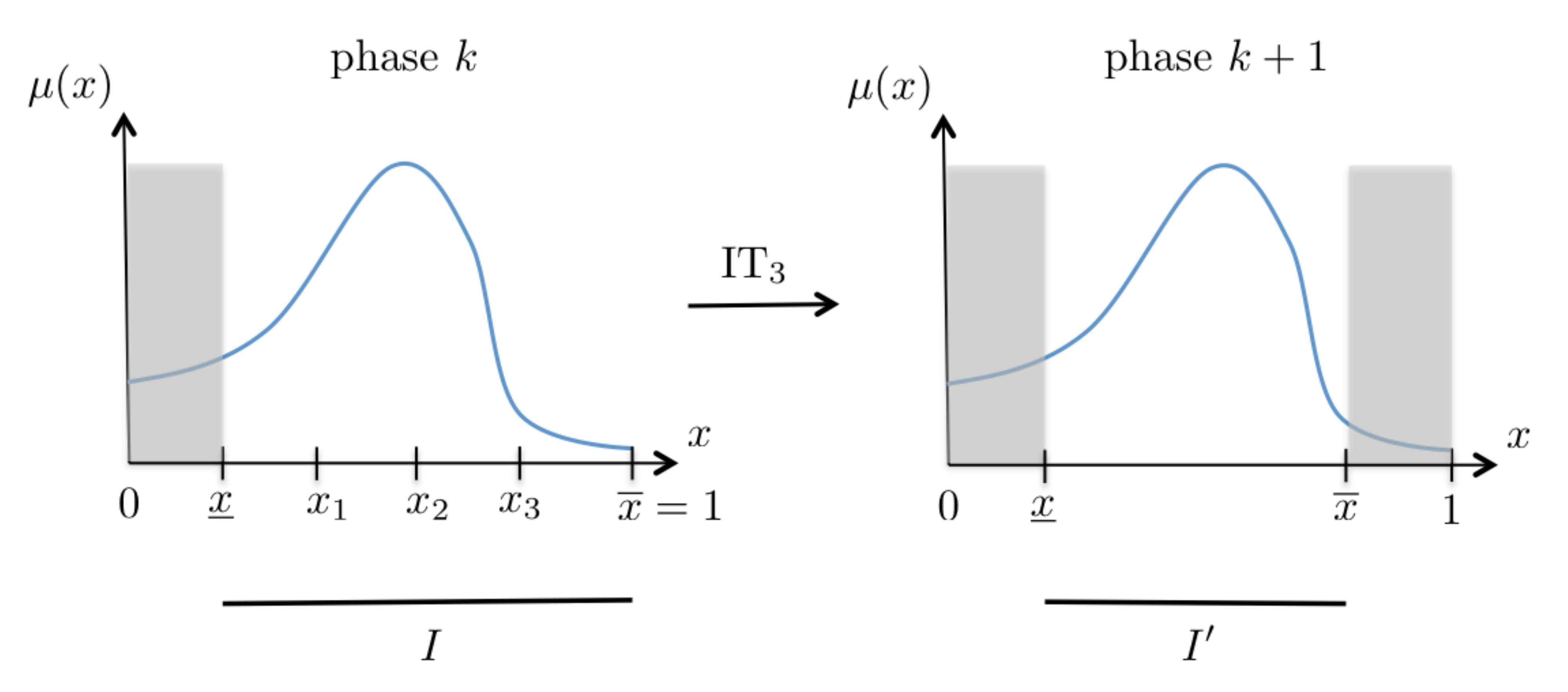}\hspace{1cm}
	\caption{A phase in the SP algorithm. In this phase, we applied the interval narrowing subroutine IT${}_3$ ($K=3$). The shaded area corresponds to parts of the interval $[0,1]$ that do not contain the best arm with high probability.}
	\label{fig:pic1}
\end{center}
\end{figure}

We present here a family of sequential arm selection rules, referred to as Stochastic Polychotomy (SP). These algorithms consist in successively narrowing an interval in $[0,1]$ while ensuring that the best arm $x^\star$ remains in this interval with high probability. Under the SP algorithms, the set of rounds is divided into \textit{phases}, where each phase consists in running a subroutine narrowing the interval containing the best arm. The narrowing subroutine used the SP algorithms, and referred to as IT${}_K$ (Interval Trimming with $K$ sampled arms), starts with an interval $I=[\underline{x},\overline{x}]$ and $K$ arms $x_1,\ldots,x_K$ with $\underline{x}\le x_1< \ldots < x_K\le \overline{x}$. It samples these $K$ arms until a decision is taken to reduce the interval $I$ and to output interval $I'$ equal to either $I_1=[\underline{x},\max\{x_k: x_k<\overline{x}\}]$ or $I_2=[\min\{x_k:x_k>\underline{x}\},\overline{x}]$. The subroutine IT${}_K$ is described in details in the next subsection, and its outcome is illustrated in Figure \ref{fig:pic1}. 

The pseudo-code of the Stochastic Pentachotomy algorithm, an example of SP algorithm, is presented in Algorithm \ref{alg:sd}. It uses the narrowing subroutine IT$_3$ exploiting samples from three arms in the interior of the input interval. IT$_3$ splits the input interval into five parts (hence the name "Pentachotomy"), and outputs a trimmed interval (referred to as $I'$ in the pseudo-code) and its running time (expressed in number of rounds, and referred to as $\ell$ in the pseudo-code). The subroutine IT$_3$ takes as input an interval, a time horizon (equal to the remaining number of rounds in the bandit problem), as well as a parameter controlling its risk, defined as the probability that the subroutine outputs an interval that does not contain the arm with the highest reward. In the Stochastic Pentachotomy algorithm, the risk parameter in IT$_3$ is always taken equal to $T^{-\gamma}$ where $\gamma>1/2$. This choice will ensure that the regret of the algorithm has an optimal scaling in $T$. Note that LSE, the stochastic version of Golden section search algorithm, belongs to the family of SP algorithms (for LSE, $K=4$, $x_1=\underline{x}$, and $x_4=\overline{x}$). 
 
\begin{algorithm}[t!]
   \caption{The Stochastic Pentachotomy algorithm}
   \label{alg:sd}
\begin{algorithmic}
\STATE {Input parameters:} time horizon $T$ and confidence parameter $\gamma > 1/2$.
\STATE {Initialization:} $I\leftarrow [0,1]$ and $s\leftarrow T$.
\STATE {While $s>0$:}
\STATE \hspace{1cm} Run IT$_3(I,s,T^{-\gamma})$ and let $(I',\ell)$ be its output,
\STATE \hspace{1cm} $I\leftarrow I'$,
\STATE \hspace{1cm} $s\leftarrow s-\ell$.
\end{algorithmic}
\end{algorithm}

\subsection{\texorpdfstring{IT${}_K$}{ITK}: Asymptotically Optimal Sequential Tests for Interval Trimming}

The narrowing subroutines used in each phase of SP algorithms can be interpreted as sequential tests whose final decision is to trim a specific part of the input interval. The IT${}_K$ subroutine belongs to the following generic family ${\cal T}$ of sequential tests. 

\noindent
\textbf{Sequential Tests for Interval Trimming.} A sequential test $\chi\in {\cal T}$ takes as inputs (i) an interval $I=[\underline{x}, \overline{x}] \subset [0,1]$, and $K$ arms to sample from $x_1,\ldots,x_K$ with $\underline{x}\le x_1< \ldots < x_K\le \overline{x}$, and (ii) a time horizon $s$ that represents the maximum number of samples the test can gather. In round $n\le s$, the sequential test decides either to terminate and to output a reduced interval $I_1=[\underline{x},\max\{x_k: x_k<\overline{x}\}]$ or $I_2=[\min\{x_k:x_k>\underline{x}\},\overline{x}]$, or to acquire a new sample from one of the arms $x_1,\ldots,x_K$. The successive decisions taken under sequential test $\chi$ are represented by $S^\chi(n)\in \{0,1,2\}$. For $n\le s$, if $S^\chi(n) = 1$, the sequential test terminates and outputs the interval $I_1$. Similarly if $S^\chi(n) = 2$, $\chi$ terminates and outputs $I_2$. When $S^\chi(n)=0$ and $n<s$, the sequential test further samples an arm $x^\chi(n)$ in $\{ x_1,\ldots,x_K\}$. Finally, if $S^\chi(s)=0$, we say that the test does not terminate, and it outputs the initial interval $I$. The sequential test is adapted in the sense that $x^\chi(n)$ and $S^\chi(n)$ are ${\cal F}_{n-1}^\chi$-measurable. We denote by $S^\chi\in \{0,1,2\}$ the final outcome of the test $\chi$. The length of a sequential test $\chi$ is defined as $L^\chi = \inf\{n \leq s: S^\chi(n)\neq 0\}$ if the test terminates and $L^\chi = s$ otherwise. $\chi$ also outputs its length.

\medskip
\noindent
\textbf{IT${}_K$ Subroutine.} To specify our sequential test $\chi=$IT${}_K$, we introduce the following notation. Define the sets of functions $B_u = \{ \mu \in {\cal U} : x^\star \notin I_u \}$, $u \in \{1,2\}$. We also introduce for any $u\in \{1,2\}$, the function $i_u: \mathbb{R}_+^K\to \mathbb{R}$ with
$$
i_u( \mu_1,\dots,\mu_K ) = \inf_{\lambda \in B_u} \sum_{k=1}^K \KL( \mu_k,\lambda(x_k)).
$$
We further denote by $t^\chi_k(n) = \sum_{n'=1}^{n\vee L^\chi} \indic\{ x^\chi(n') = x_k \}$ the number of times arm $x_k$ is sampled up to time $n$ and before the test $\chi$ terminates. Finally, we define the empirical average reward of arm $x_k$ up to round $n\le L_\chi$ as:
$$
\hat\mu_k(n) = \frac{1}{t_k^\chi(n)} \sum_{n'=1}^n  X_{n'}( x_k )  \indic\{ x^\chi(n) = x_k \},
$$
if $t_k^\chi(n) > 0$ and $ \hat\mu_k(n)= 0 $ otherwise. Let $\hat{\mu}(n)=(\hat{\mu}_1(n),\ldots,\hat{\mu}_K(n))$ and $\bar{t}^\chi(n) = \min_{1 \leq k \leq K} t_k^\chi(n)$. $\chi=$IT${}_K$ samples $K$ arms in the interior of $I=[\underline{x},\overline{x}]$, i.e., $\underline{x}<x_1<\ldots <x_K<\overline{x}$. To simplify the presentation, we assume that for $k=1,\ldots,K$, $x_k =  \underline{x} + k (\overline{x} -\underline{x})/(K+1)$. This assumption is not crucial, and our analysis remains valid for any choice of arms provided that they lie in the interior of $I$. 

The sequential test $\chi=$IT${}_K$ has inputs $I$ and $s$, as any other test in ${\cal T}$. However $\chi$ takes an additional input $\zeta>0$, used to control its risk. Now IT$_K(I,s,\zeta)$ is defined as follows. 

Define 
\eqs{
F(f,s,K) = e^{K + 1  - f} ( f \ceil{ f \log(s) }/K )^K,
} 
and let $f(s,\zeta) \geq K+1$ be such that $F(f(s,\zeta),s,K) \leq \zeta$ (the precise choice of $f(s,\zeta)$ is free). The test proceeds as follows:
For any $n\leq s$:
\begin{itemize}
\item[(i)] If there exists $u\in \{1,2\}$ such that $\bar{t}^\chi(n) i_u(\hat\mu(n) ) \geq f(s,\zeta)$, then $S^\chi(n) = u$, i.e., $\chi$ terminates and  its final output is $S^\chi = u$ (ties are broken arbitrarily if both conditions $\bar{t}^\chi(n) i_u(\hat\mu(n) ) \geq f(s,\zeta)$ for $u=1,2$ hold).
\item[(ii)] Otherwise $S^\chi(n) = 0$, and $\chi$ samples arm $x^\chi(n) = x_{1+ (n \hbox{ mod }K)}$.
\end{itemize}
The sequential test $\chi$ outputs the interval $I_{S^\chi}$ where $I_0=[\underline{x},\overline{x}]$, $I_1=[\underline{x},x_{K}]$ and $I_2=[x_1,\overline{x}]$, and its length $L^\chi$.

The performance (i.e. the minimax risk and length) of IT$_{K}$ will be analysed in Section \ref{sec:perf}. In view of the results derived in Sections \ref{sec:perf} and \ref{sec:low}, IT$_{K}$ is asymptotically optimal among the sequential tests in ${\cal T}$. The design of IT$_{K}$ (e.g. the use of functions $i_u$, $u\in \{1,2\}$) is actually motivated by the fundamental performance limits of tests in ${\cal T}$ derived in Section~\ref{sec:low}.
\begin{remark}\label{rem:fbar}
In the following sections, we will mainly consider the case where the risk $\zeta = s^{-\gamma}$ with $\gamma > 0$. In this case, one may choose $f(s,s^{-\gamma}) = \overline{f}(s) := \gamma \log(s) + 3 K \log(\log(s)) + C$, where $C > 0$ is independent of $s$ and $\gamma$. 
\end{remark}

 \subsection{\texorpdfstring{IT$_3'$}{IT3'}: A Computationally Efficient Sequential Test}
 
Next we present IT$_3'$, a  sequential test which is computationally simpler than IT$_3$. IT$_3'$ is not asymptotically optimal, but its implementation is much simpler than that of IT$_3$. Its rationale involves calculating an explicit lower bound of functions $i_u$, $u\in \{ 1,2\}$, and hence IT$_3'$ does not require us to compute $i_u$. For $\epsilon \geq 0$, we define the function KL$^{\star,\epsilon}: \RR^2 \to \RR$ as:
\eqs{
\text{KL}^{\star,\epsilon}( \mu_1,\mu_2) = \indic \{ \mu_1 < \mu_2    \} \left[ \KL \left(\mu_1 + \epsilon, \frac{\mu_1 +  \mu_2}{2} - \epsilon \right) + \KL \left( \mu_2 - \epsilon, \frac{\mu_1 +  \mu_2}{2} + \epsilon \right) \right].
}
and $\text{KL}^{\star}( \mu_1,\mu_2) = \text{KL}^{\star,0}( \mu_1,\mu_2)$. The sequential test $\chi'=$ IT$_3'$ with inputs $I$, $s$ and $\zeta$ is defined by: for any $n\leq s$, 
\begin{itemize}
\item[(i)] If $\bar{t}^{\chi'}(n) \text{KL}^{\star}( \hat\mu_1(n),\hat\mu_2(n)) \geq f(s,\zeta)$, then $S^{\chi'}(n) = 1$, i.e., $\chi'$ terminates and  its final output is $S^{\chi'} = 1$. Similarly if $\bar{t}^{\chi'}(n) \text{KL}^{\star}( \hat\mu_3(n),\hat\mu_2(n)) \geq f(s,\zeta)$, then $S^{\chi'}(n) = 2$.
\item[(ii)] Otherwise $S^{\chi'}(n) = 0$, and $\chi'$ samples arm $x^{\chi'}(n) = x_{1+ (n \hbox{ mod }3)}$.
\end{itemize}

\section{Performance Analysis of the Stochastic Pentachotomy Algorithm}\label{sec:perf}

In this section, we analyze the performance of the Stochastic Pentachotomy algorithm. To this aim, we first study how the interval trimming subroutines IT${}_K$ (for $K\ge 3$) and IT${}_3'$ perform.

\subsection{Minimax Risk and Length of \texorpdfstring{IT${}_K$}{ITK}}

Let $\chi\in {\cal T}$ be a sequential test for interval trimming. For any $\mu\in {\cal U}$, the risk $\alpha^\chi( \mu )$ of $\chi$ is the probability that $\chi$ outputs an interval that does not contain the optimal arm, i.e, $\alpha^\chi( \mu ) = \sum_{u=1}^2  \indic \{ \mu \in B_u \} \PP_{\mu}[ S^\chi = u ]$. The minimax risk of $\chi$ is then defined as $\alpha^\chi= \sup_{ \mu \in {\cal U} } \alpha^\chi( \mu )$. Observe that a test that does not terminate (almost surely) has a risk equal to 0, but then its length would be maximal. The analysis of the performance of a test hence consists in characterizing the trade-off between its risk and its length. The next theorem provides upper bounds of the minimax risk of IT$_K$, as well as of the number of times arms are sampled before the test terminates.

  
\begin{theorem}\label{th:elim}
Let $K\ge 3$ and $I \subset [0,1]$.\\
(i) For any $s\ge 1$, the minimax risk of IT$_K(I,s,\zeta)$ is smaller than $\zeta$. \\
(ii) Let $\gamma > 0$, $u \in \{1,2\}$ and $k \in 1,\ldots,K$. For all $\mu \in {\cal U} \setminus B_u$, the test $\chi=$IT$_K(I,s,s^{-\gamma})$ satisfies:
	\eqs{
		\lim\sup_{s \to \infty} \frac{ \EE_{\mu}[ t_k^\chi(s) ]}{\log(s)} \leq \frac{\gamma}{i_u(\mu(x_1),\dots,\mu(x_K))}. 
	}
\end{theorem}

The above theorem provides asymptotic guarantees on the length of IT$_K$. Next, we provide a finite-time analysis of the length of IT$_3$ and IT$_3'$, and we also derive an upper bound of the minimax risk of IT$_3'$.

\medskip
\noindent
\textbf{Finite-time analysis of IT$_3$ and IT$_3'$.} The next theorem provides explicit upper bounds on the expected length of IT$_3$ and IT$_3'$. A high-probability upper-bound on the test length is also provided. This result relies on an explicit lower bound of $i_u(\mu_1,\mu_2,\mu_3)$. Theorem~\ref{th:IT3} will be instrumental in the regret analysis of the Stochastic Pentachotomy algorithm. We restrict the analysis to Bernoulli rewards. This is mainly for simplicity, and the proof techniques can be extended to sub-Gaussian rewards with straightforward modifications.

\begin{theorem}\label{th:IT3}
Consider $I \subset [0,1]$, $\gamma > 0$ and tests $\chi \in \{ \text{IT}_3(I,s,s^{-\gamma}), \text{IT}_3'(I,s,s^{-\gamma}) \}$. \\
(i) $\chi$ has minimax risk less than $s^{-\gamma}$. \\
(ii) Define $m=1$ if $x^\star \in [x_2,\overline{x}]$ and $m=3$ otherwise. Define $\delta = (\mu(x_2) - \mu(x_m))/2$. Then, we have that for all  $0 < \epsilon < \delta/2$, for all $k=1,2,3$ and all $s\ge 1$:
	\eqs{
		\EE_{\mu}[ t_k^{\chi}(s) ] \leq \frac{\overline{f}(s)}{\text{KL}^{\star,\epsilon}( \mu(x_m),\mu(x_2)) } + 2 \epsilon^{-2}.
	}
(iii) We have the following inequalities:
\als{
	\text{(a) } & \PP_{\mu}[t_k^{\chi}(s) \geq  8 \overline{f}(s) \delta^{-2}  ] \leq 2 e^{-\overline{f}(s)} \sk
	\text{(b) } & \EE_{\mu}[ t_k^{\chi}(s) ] \leq \frac{32 + \overline{f}(s)}{\delta^2} \sk 
	\text{(c) } & \lim\sup_{s \to \infty}	\frac{ \EE_{\mu}[ t_k^{\chi}(s) ]}{\log(s) } \leq \frac{\gamma}{\text{KL}^{\star}( \mu(x_m),\mu(x_2)) }.
	}
\end{theorem} 

Recall that $\overline{f}(s) := \gamma \log(s) + 9 \log(\log(s)) + C$, see Remark 1.

\subsection{Regret Upper Bounds of the SP algorithm}

Next, we analyze the regret of the Stochastic Pentachotomy algorithm. We refer to as \ouralgoprime the algorithm using the narrowing subroutines IT$_3'$ (instead of IT$_3$ for \ouralgom). Recall that the successive narrowing subroutines IT$_3$, the risk is always chosen equal to $T^{-\gamma}$, as specified in Algorithm \ref{alg:sd}. We first derive an upper bound valid for all $\mu\in {\cal U}$ and all time horizon $T$. We then specify the bound when $\mu$ behaves as $\mu(x)=\mu(x^\star)-C|x-x^\star|^\xi$ locally around its maximizer $x^\star$ for some $\xi, C>0$. To simplify the presentation, our bounds are stated and proved for Bernoulli rewards, but the analysis can be extended to other exponential families of distributions. 

Let $\mu\in {\cal U}$. For any $\Delta>0$, define the following functions, which will be used to state our regret upper bound:
\als{
g_\mu(\Delta) &= \mu^\star - \max( \mu( x^\star - \Delta ) , \mu( x^\star + \Delta ) ) \sk
h_\mu(\Delta) &=  \min \left\{    \min_{ x \in [ x^\star , x^\star + \Delta/4 ]} ( \mu(x) - \mu(x + \Delta/4)),   \min_{ x \in [ x^\star- \Delta/4, x^\star]} ( \mu(x) - \mu(x - \Delta/4)) \right\}  
}

\begin{theorem}\label{th:regret}
Let $\psi = 3/4$. Under Algorithm $\pi=$ \ouralgo or $\pi=$ \ouralgoprimem, for all $\mu \in {\cal U}$, all $T\ge 1$, and all $N\ge 1$, the regret satisfies:
\eqs{
R^\pi(T) \leq  \mu^\star N T^{1-\gamma} + T g_\mu( \psi^{N}) + 3 (\overline{f}(T) + 32) \sum_{N'=0}^{N-1} g_\mu( \psi^{N'} ) h_\mu( \psi^{N'} )^{-2}.
}
\end{theorem}

We now make the regret upper bound of Theorem~\ref{th:regret} explicit by considering a particular class of unimodal functions. 

\begin{definition}\label{def:mu} 
For given $0 < C_1 \leq C_2 < \infty$, we define ${\cal U}(C_1,C_2)$ the set of all unimodal functions $\mu \in {\cal U}$ for which there exists $\xi > 0$ such that:

(P1) $\mu(x) - \mu(y) \geq C_1 ( |x^\star - y|^{\xi} - |x^\star - x|^{\xi} )$ for all $0 \leq y \leq x \leq x^\star$ and $x^\star \leq x \leq y \leq 1$. \\
 
(P2) $|\mu^\star - \mu(x)| \leq C_2|x^\star - x|^{\xi}$ for all $x \in [0,1]$.
\end{definition}

Note that for any $\mu \in {\cal U}$ such that $|\mu^\star - \mu(x)| \sim_{x \to x^\star} C |x^\star - x|^{\xi}$ with $C > 0$, there exists $C_1 > 0$ suitably small and $C_2 < \infty$ suitably large such that $\mu \in {\cal U}(C_1,C_2)$. Also note that if $\mu\in{\cal U}$ is differentiable on $[0,1] \setminus \{x^\star \}$, with $C_{1} |x^\star - x |^{\xi-1} \leq  | \mu'(x) | \leq  C_{2} |x^\star - x |^{\xi-1}$, then $\mu \in {\cal U}(C_1,C_2)$. 
\begin{theorem}\label{th:regretclass}
Assume that the algorithm $\pi=$ \ouralgo or $\pi=$ \ouralgoprimem is parametrized by $\gamma > 1/2$. For all $\mu \in {\cal U}(C_1,C_2)$, the regret satisfies:
\eqs{
	R^\pi(T) \leq \frac{2 \psi^{-3\xi/2} C_2}{C_1 a_{\xi}} \sqrt{ \frac{3 T (\overline{f}(T) + 32)}{\psi^{-\xi} - 1} } +  \mu^\star T^{1-\gamma} \frac{\log( T C_1 \psi^{-\xi})}{\xi \log( 1/\psi )}   =  O( \sqrt{T \log(T)}).
}
where $a_{\xi} = 4^{-\xi} \min( 1 , 2^{\xi} - 1)$, and where $\xi$ is the parameter associated with $\mu$ in Definition \ref{def:mu}.

\end{theorem}
Theorem~\ref{th:regretclass} states that \ouralgo and \ouralgoprime are order-optimal for all reward functions in ${\cal U}(C_1,C_2)$ (with arbitrary $C_1$ and $C_2$). They achieve a regret scaling as $O( \sqrt{T \log(T)})$ without the knowledge of the behaviour of the reward function around its maximizer. Although the regret upper bound of Theorem~\ref{th:regretclass} is stated for reward functions in class ${\cal U}(C_1,C_2)$, we emphasize again that $C_1$, $C_2$ and $\xi$ are not input parameters of the algorithms.


\subsection{Optimization error of the SP algorithm}\label{subsec:optimizationerror}
We conclude this section by deriving an upper bound on the optimization error of algorithms SP and SP'. 

\begin{theorem}\label{th:optimerror}
Let $\psi=3/4$. Assume that the algorithm $\pi=$ \ouralgo or $\pi=$ \ouralgoprimem is parametrized by $\gamma > 1/2$. For all $\mu \in {\cal U}(C_1,C_2)$, the optimization error under $\pi$ satisfies:
\eqs{
E^{\pi}(T) \leq \frac{C_2}{C_1 a_{\xi}} \sqrt{ \frac{24 \overline{f}(T)}{T(\psi^{-2\xi}-1)}} +  \frac{3 T^{-\gamma} \mu^\star \log(T C_1 \psi^{-\xi})}{ \xi \log(1/\psi)} = O(\sqrt{\log(T)/T}),
} 
with $a_{\xi} = 4^{-\xi} \min( 1 , 2^{\xi} - 1)$, and where $\xi$ is the parameter associated with $\mu$ in Definition \ref{def:mu}.
\end{theorem}

\section{Fundamental Performance Limits for Interval Trimming Subroutines}\label{sec:low}
  
The next theorem provides a lower bound on the expected number of times each arm $x_k, k=1,\dots,K$ must be sampled under \emph{any} sequential test with given minimax risk. The lower bound is valid for any time horizon $s$, which contrasts with the asymptotic lower bounds usually derived in the bandit literature (see e.g. \cite{lai1985}). The proof of this lower bound relies on an elegant information-theoretic argument that exploits the log-sum inequality to derive lower bounds of KL divergence numbers. 
 
\begin{theorem}\label{th:sequential}
Let $\chi\in {\cal T}$ be a sequential test for interval trimming with minimax risk $\alpha$. Let $\mu \in {\cal U}$, and $u \in \{1,2\}$. Let $\beta= \PP_{\mu}[ S^\chi = u ]$. If $\alpha \leq \beta$, then
	\eqs{
	\inf_{\lambda \in B_u} \sum_{k=1}^{K} \EE_{\mu}[ t_k^\chi(s) ] \KL( \mu(x_k), \lambda(x_k) ) \geq \KLtwo( \beta,\alpha). 
	}
\end{theorem} 

From the above result, we deduce Corollary \ref{cor:sequential1} stating that any sequential test with time horizon $s$ and with minimax risk $s^{-\gamma}$, for $\gamma\in (0,1]$, has a length that scales at least as $\gamma \log(s)$ as $s$ grows large. Note that the sequential tests IT$_K$ match these lower bound and are hence asymptotically optimal.

\begin{corollary}\label{cor:sequential1}
Let $\gamma\in (0,1]$, $u \in \{1,2\}$, and $\mu \in {\cal U }$. Consider a sequence (indexed by $s$) of sequential tests $\chi_s$ with time horizon $s$ and minimax risk $\alpha^{\chi_s} =s^{-\gamma}$, such that  $\lim_{s\to\infty}\PP_{\mu}[S^{\chi_s} = u] = \beta > 0$. Then:
	$
	\lim\inf_{s \to \infty} \inf_{\lambda \in B_u}  \sum_{k=1}^{K} \frac{\EE_{\mu}[ t^{\chi_s}_k(s) ]}{\log(s)} \KL( \mu(x_k), \lambda(x_k) ) \geq \gamma \beta.
	$
\end{corollary}

Another consequence of Theorem \ref{th:sequential} is presented in Corollary \ref{cor:sequential2}. The latter states that it is impossible to construct a sequential test that samples at most two arms in the interior of $I$, that terminates before the time horizon $s$ with probability larger than $1/2$ and that has a minimax risk strictly less than $1/4$. Note that if a test terminates before $s$ with probability less than $1/2$, its expected length is at least $s/2$. Such a test would be useless in bandit problems since running it with time horizon $s=T$ would incur a regret linearly growing with  $T$. 

\begin{corollary}\label{cor:sequential2}
Consider the family of sequential tests running on the interval $I = [\underline{x}, \overline{x}]$, and arms $\underline{x} = x_1 < x_2 < x_3 < x_4 = \overline{x}$. There exists $\mu \in {\cal U}$, such that for any sequential test $\chi$ of this family with arbitrary finite time horizon $s$ and minimax risk $\alpha < 1/4$, we have $\PP_{\mu}[S^\chi \neq 0] \leq 1/2$ (i.e., the test does not terminate before $s$ with probability $1/2$).
\end{corollary}

Recall that Kiefer's Golden section search algorithm \cite{Kiefer1953} uses two points in the interior of the interval to reduce. Hence, the above corollary implies that it is impossible to construct a stochastic version of this algorithm that performs well without additional assumptions on the smoothness and structure of the reward function. Actually, LSE, proposed in \cite{yu2011}, is a stochastic version of the Golden section search algorithm, but to analyze its regret, additional assumptions on the structure of the reward function are made (its minimal slope and smoothness). 
 
\begin{figure}
\begin{center}
	\includegraphics[width=0.4\columnwidth]{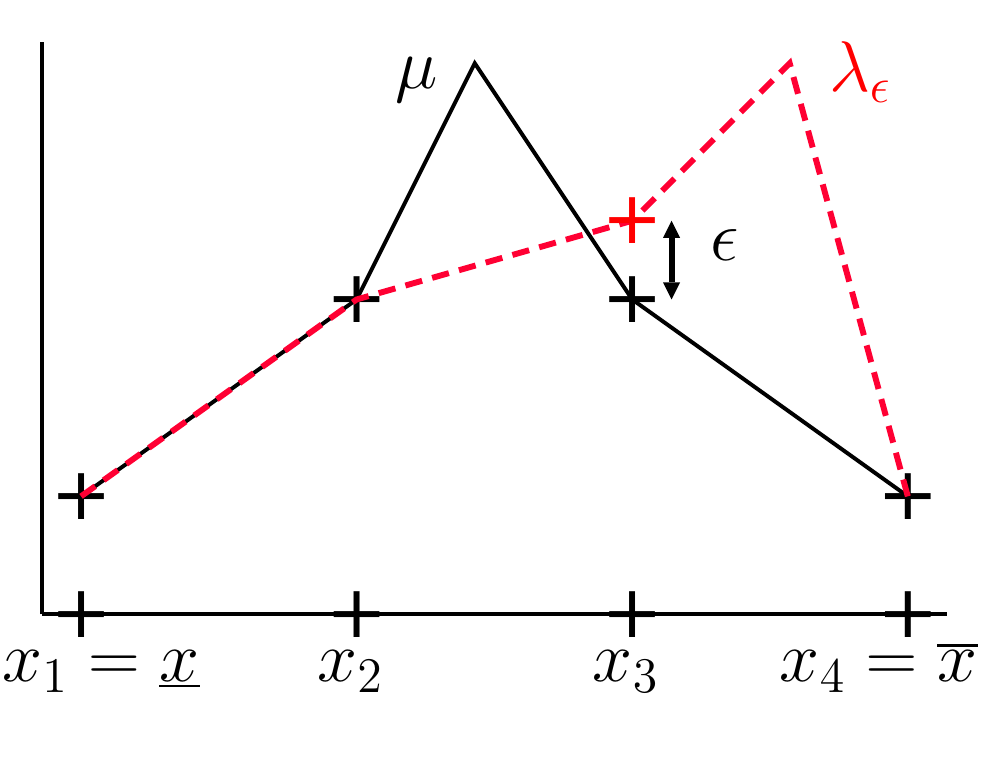}
	\caption{Illustration of Corollary~\ref{cor:sequential2}} 
	\label{fig:golden_ratio}
\end{center}
\end{figure}

Corollary~\ref{cor:sequential2} is a direct consequence of Theorem \ref{th:sequential}: the choice of the reward function $\mu$ used in Corollary \ref{cor:sequential2} is illustrated in Figure \ref{fig:golden_ratio}, and the result is obtained by considering a sequence (indexed by $\epsilon>0$) of unimodal functions $\lambda_\epsilon\in B_1$. An efficient test must distinguish between $\mu$ and $\lambda_\epsilon$ based on the reward samples at $x_1,x_2,x_3,x_4$. By letting $\epsilon \to 0$, we see that under such a test, the number of samples from $x_3$ must be arbitrary large.


\section{Numerical Experiments}\label{sec:num}

In this section, we briefly explore the performance of SP$'$ (using parameter $\gamma=0.6$), and compare it to that of two other algorithms, namely KL-UCB($\delta$) and KW. KL-UCB($\delta$) consists in applying the KL-UCB algorithm \cite{garivier2011} to the discrete set of arms $\{0,\delta,2\delta,\ldots, 1\}$. KW is the algorithm proposed in \cite{cope09}. The performance of LSE \cite{yu2011} is not reported here, since it is generally outperformed by KL-UCB($\delta$), as shown in \cite{combes2014}.

We consider two reward functions satisfying our assumptions with $\xi=1/2$ and $\xi=2$, respectively. More precisely, $\mu(x) = 1 - (2|1/2 - x|)^{\xi}$ for $x\in [0,1]$. The first function is not differentiable at its maximizer, whereas the second function is just quadratic. Note that KW should then perform well for the quadratic rewards (there the regret scales as $O(\sqrt{T})$ \cite{cope09}), but there is not guarantee that it would do well for the non-differentiable reward functions. For KL-UCB($\delta$), the optimal discretization step $\delta$ depends on the smoothness of the reward function, and is set to $(\log(T) / \sqrt{T} )^{1/\xi}$. 

\begin{figure}
\begin{center}
	\includegraphics[width=0.4\columnwidth]{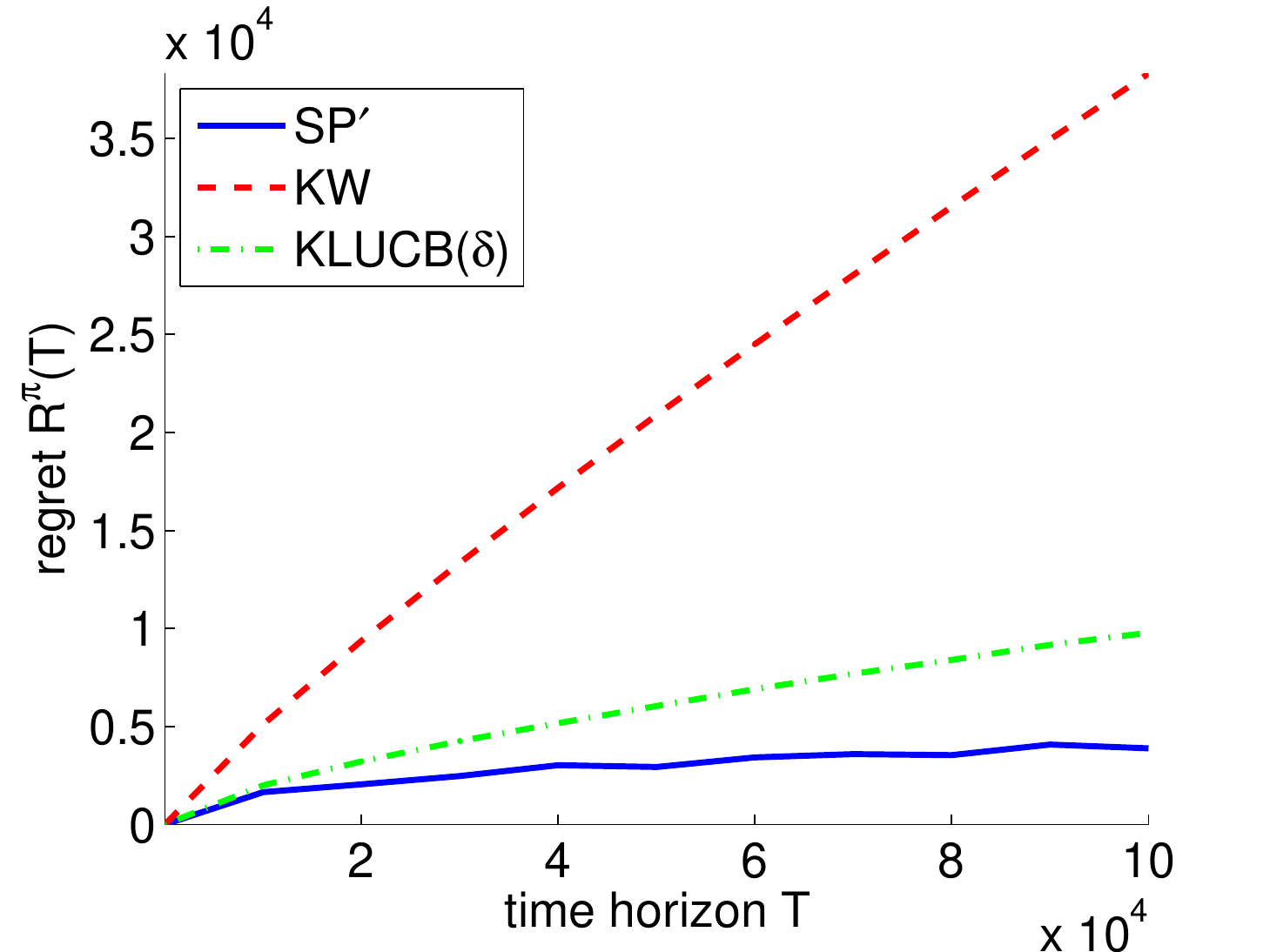}
	\includegraphics[width=0.4\columnwidth]{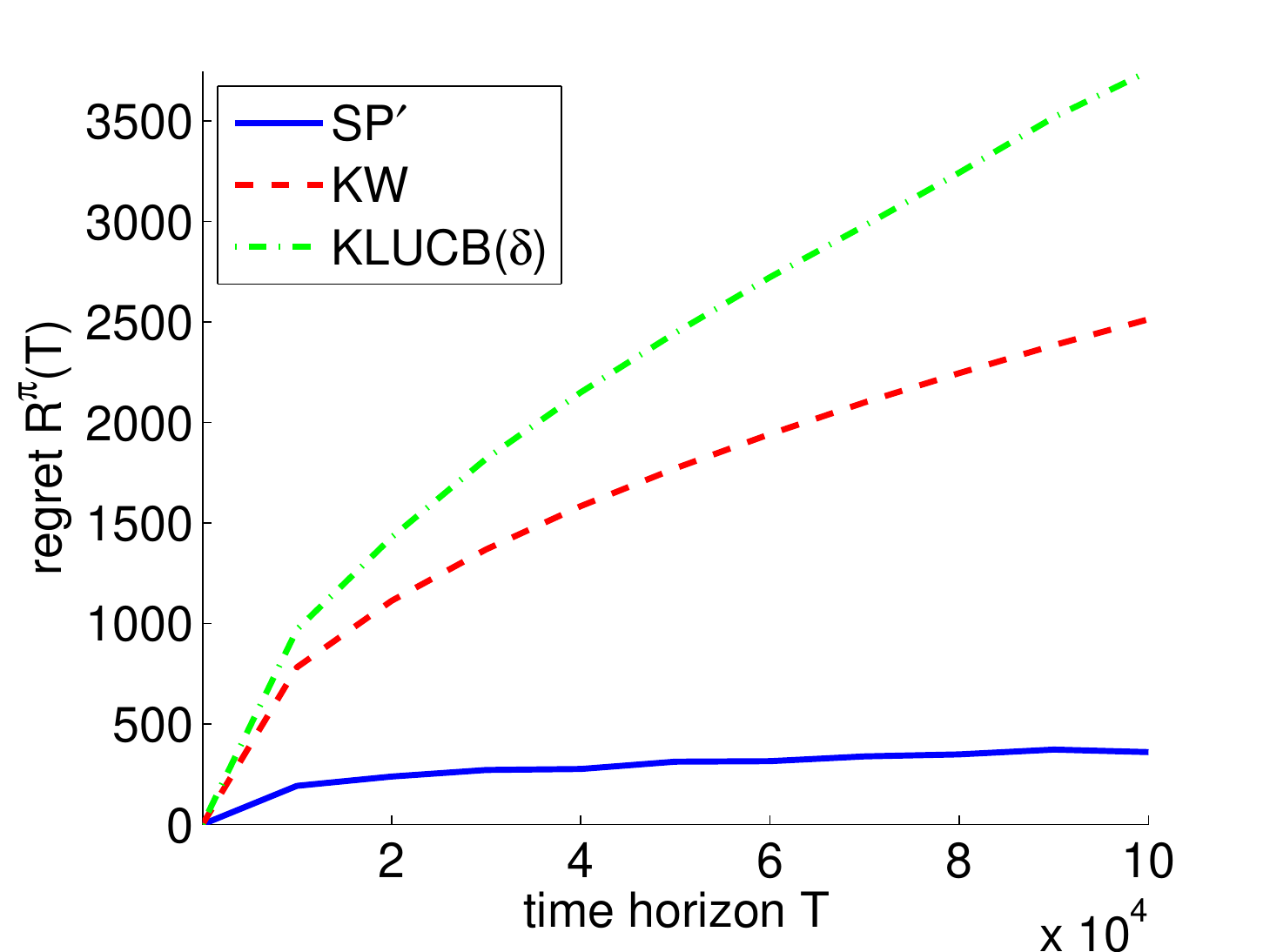}
	\caption{Regret of various algorithms for $\mu(x) = 1 - (2|1/2 - x|)^{\xi}$, $\xi = 0.5$ (left), and $\xi = 2$ (right).}
	\label{fig:regret_t}
\end{center}
\end{figure}

In Figure \ref{fig:regret_t}, we present the regret of the various algorithms (averaged over 10 independent runs). Observe that without the knowledge of the smoothness of the function, SP$'$ is able to significantly outperform the two other algorithms. As expected, KW does not perform well when $\xi=1/2$, but outperforms KL-UCB$(\delta)$ for $\xi=2$. 
    
\begin{figure}[ht]
\begin{center}
	\includegraphics[width=0.45\columnwidth]{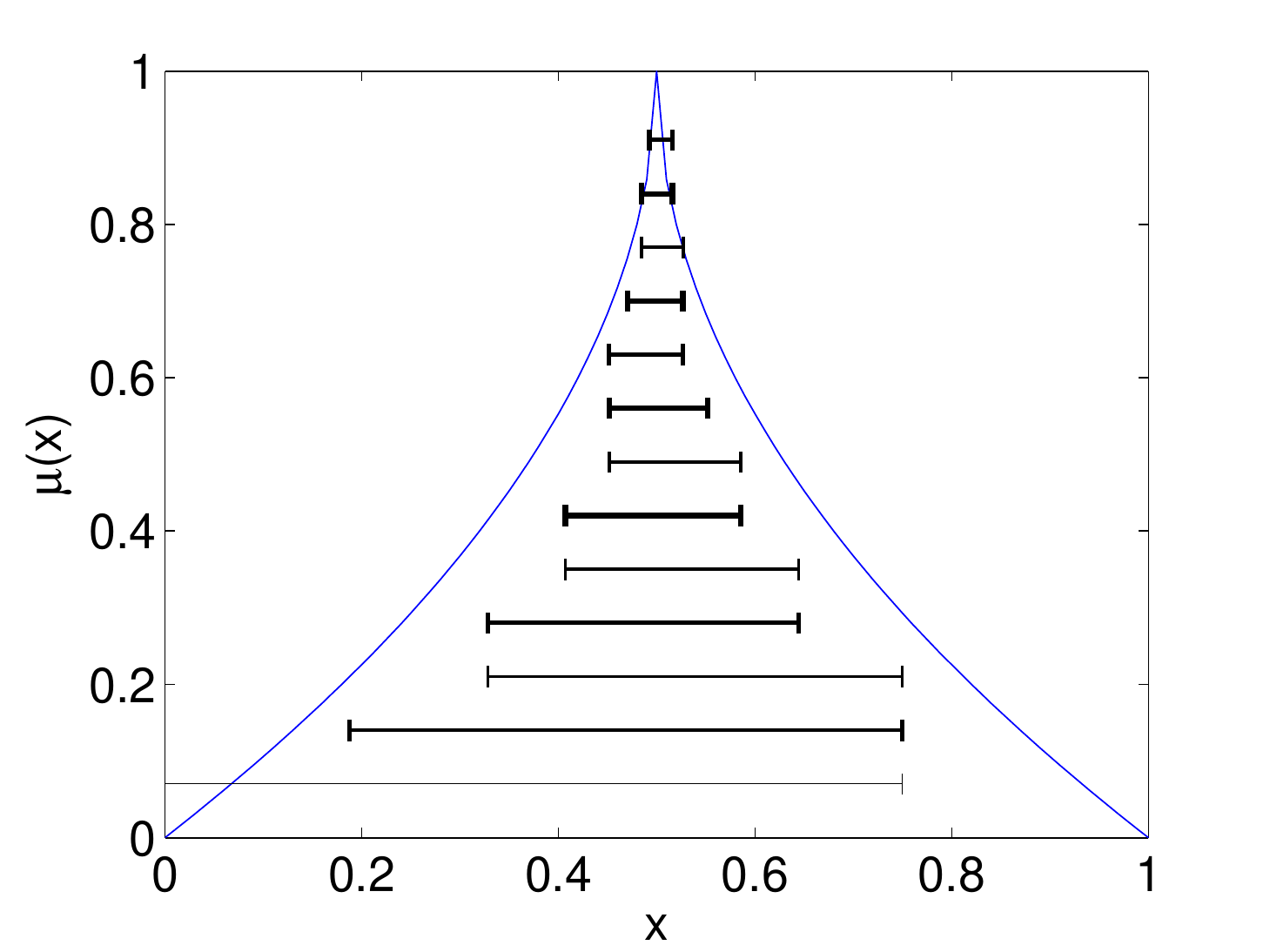}
	\includegraphics[width=0.45\columnwidth]{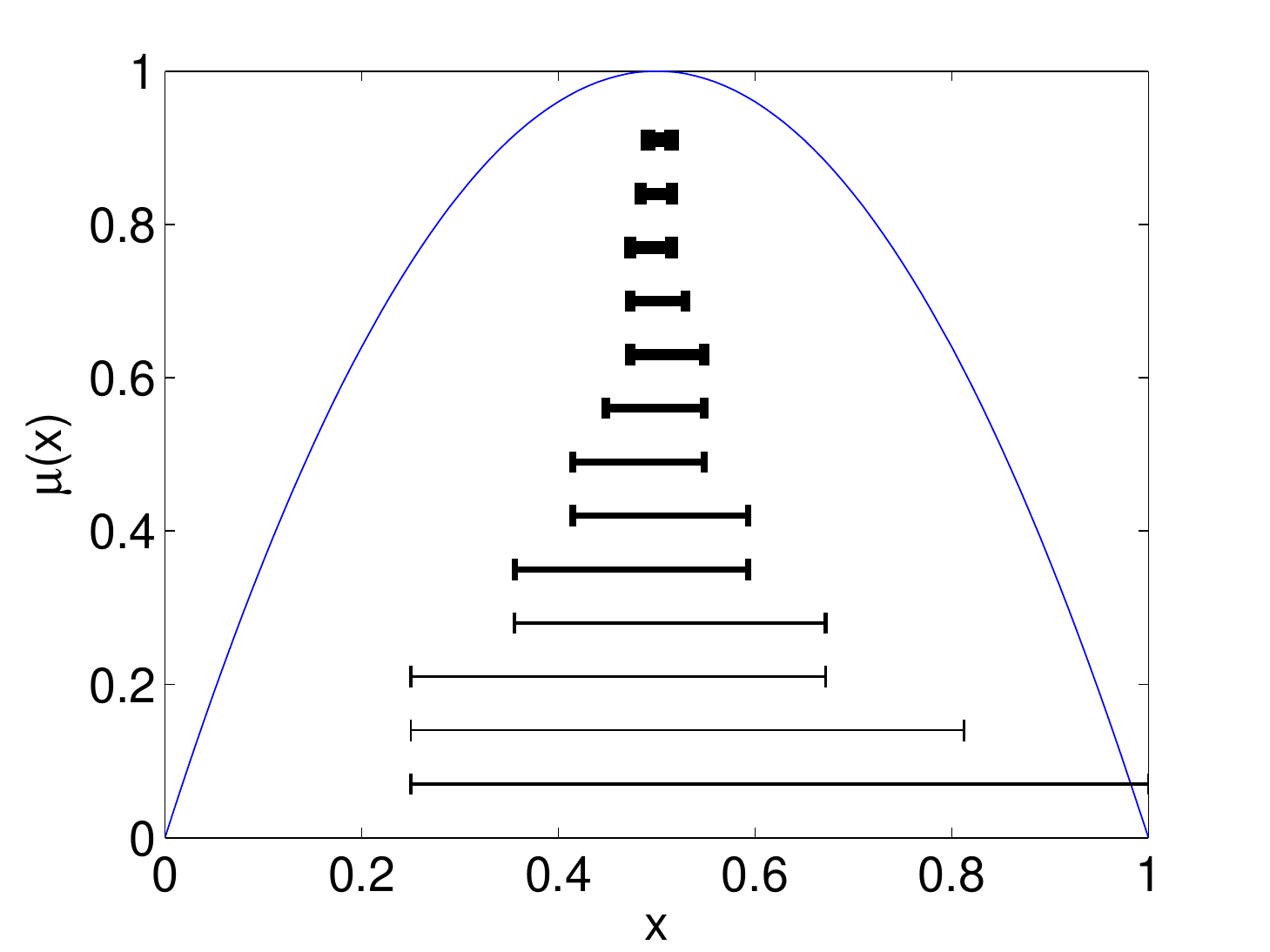}
	\caption{Illustration of a run of SP$'$ with reward function $\mu(x) = 1 - (2|1/2 - x|)^{\xi}$, $\xi = 0.5$ (left), and $\xi = 2$ (right) and time horizon $T=10^{6}$.}
	\label{fig:sd_visual}
\end{center}
\end{figure}

Figure~\ref{fig:sd_visual} presents a graphical illustration of a typical run of SP$'$ with reward function $\mu(x) = 1 - (2|1/2 - x|)^{\xi}$, $\xi = 0.5$ (left), and $\xi = 2$ (right), time horizon $T = 10^6$ and $\gamma = 0.6$. We represent the shape of $\mu$ and the successive intervals returned by IT$_3'$, starting at the bottom of the y-axis. The thickness of the segments is an increasing function of the length of IT$_3'$. In both cases, we observe that the successive intervals contain the optimal arm $x^\star$. When the search interval gets narrower (we are closer to the peak), the intervals get thicker since the duration of the test increases when the separation between arms $\{x_1,x_2,x_3\}$ decreases. Also remark that when the expected reward function is flatter (here $\xi=2$), the algorithm tends to spend more time on each given interval. Additional numerical experiments are presented in Appendix. 

\section{Conclusion} 

In this paper, we have presented the first order-optimal algorithms for one-dimensional continuous unimodal bandit problems that do not explicitly take into account the structure or the smoothness of the expected reward function. In some sense, the proposed algorithm learns and adapts its sequential decisions to the smoothness of the function. Future work will be devoted to applying the techniques used to devise our algorithms to other structured bandits with continuum set of arms (i.e., Lipschitz or convex bandits). We also would like to extend our analysis to the case where the set of arms lies in a space of higher dimension.
\clearpage
\bibliography{RA}
\bibliographystyle{abbrv}
\newpage

\appendix

\section{Additional numerical experiments}\label{sec:numsup}

\begin{figure}[ht]
\begin{center}
	\includegraphics[width=0.45\columnwidth]{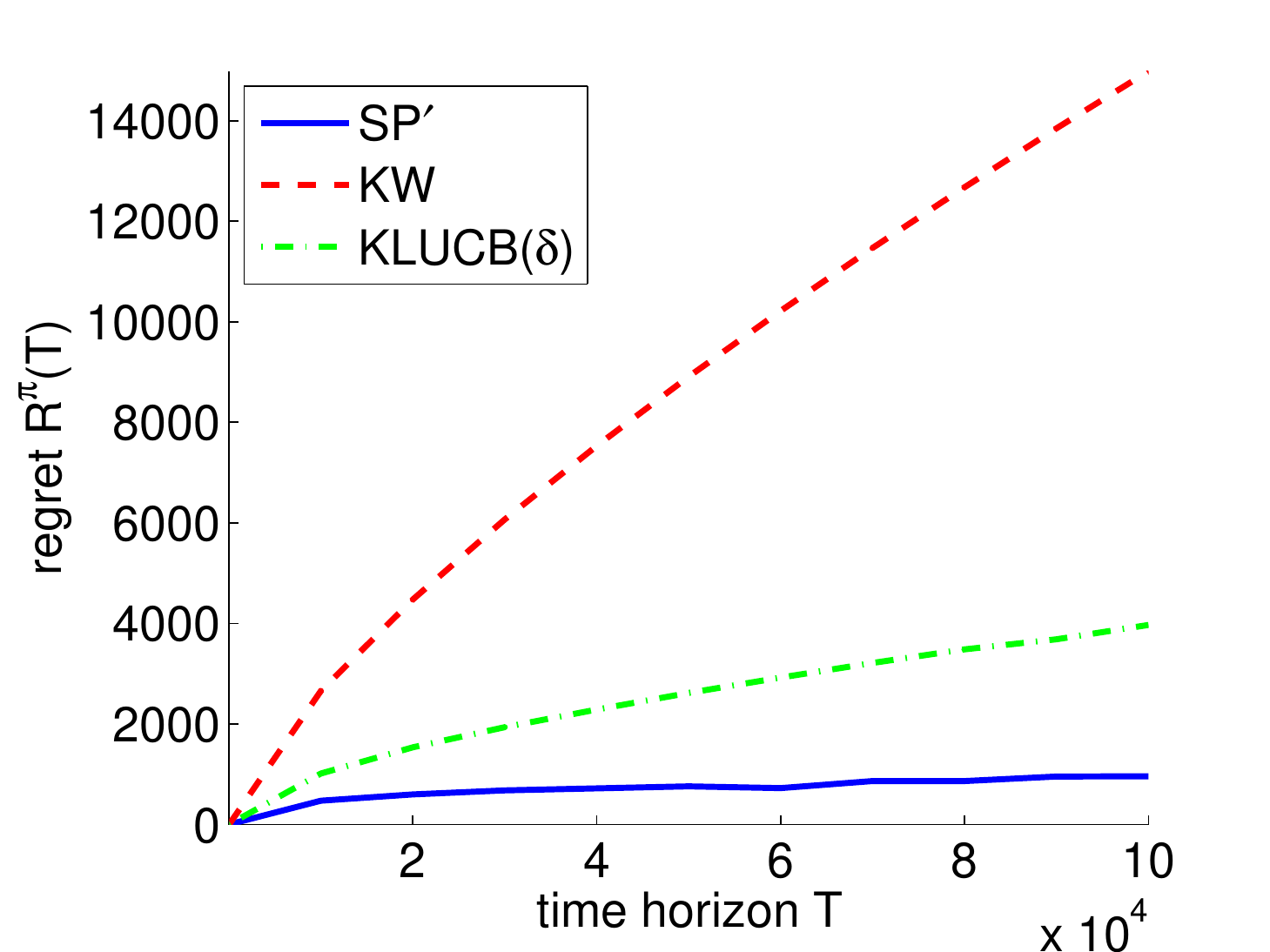}
	\includegraphics[width=0.45\columnwidth]{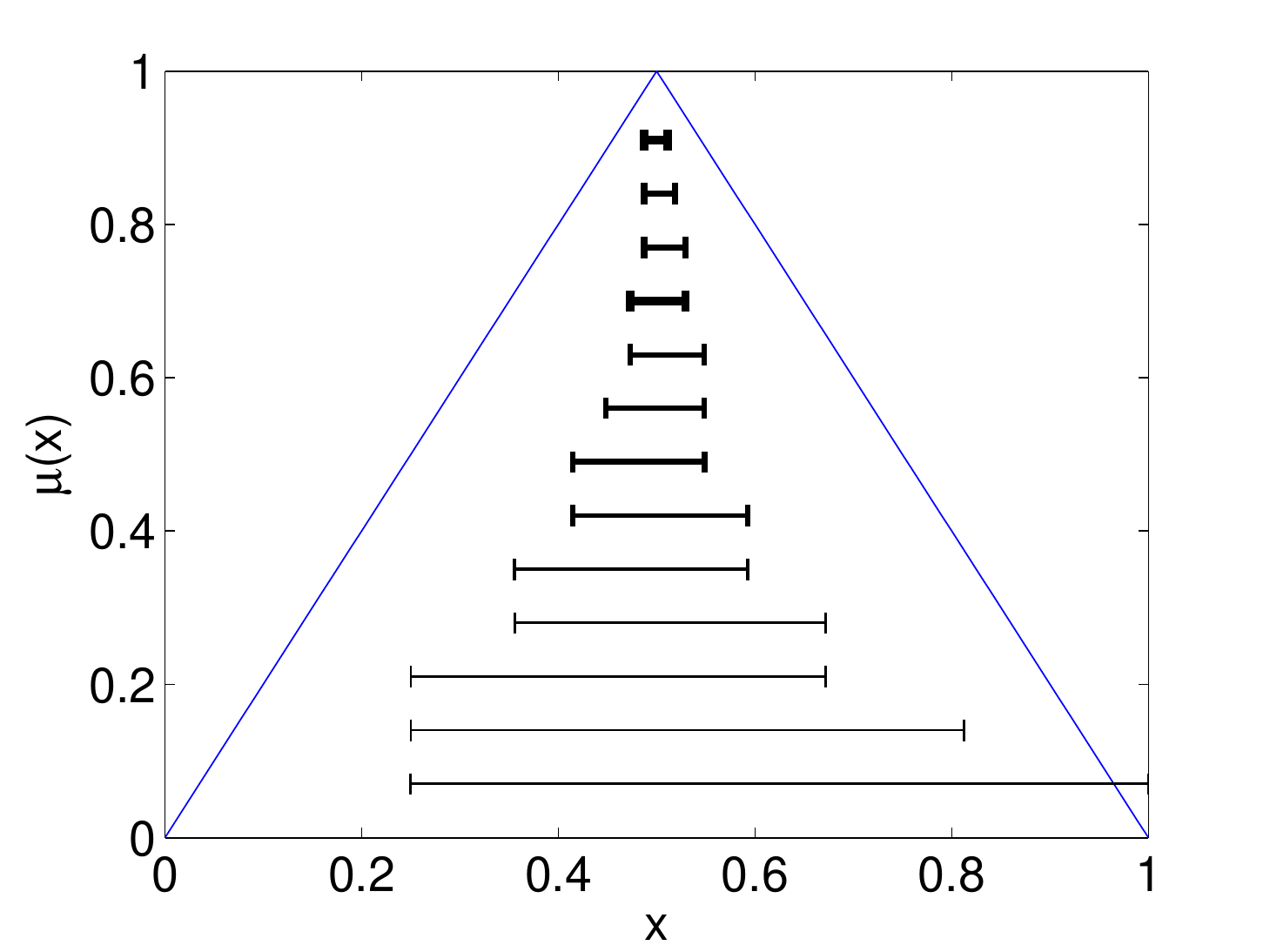}
	\caption{Reward function: $\mu(x)= 1 - (2|1/2 - x|)$. (Left) Regret vs time of various algorithms. (Right) Illustration of a run of SP$'$ with time horizon $T=10^{6}$.}
	\label{fig:111}
\end{center}
\end{figure}

Figure \ref{fig:111} compares the regret of the various algorithms for a triangular reward function  $\mu(x)= 1 - (2|1/2 - x|)$,  and illustrates a typical run of the SP$'$ algorithm for such a reward function with time horizon $T = 10^6$ and $\gamma = 0.6$.

\section{Proofs}\label{sec:proofs}

\subsection{Proof of Theorem \ref{th:elim}}

\underline{Proof of (i) (Minimax risk).}
Let $\mu\in {\cal U}$, and consider the test $\chi=$IT$_K$. By definition, its risk is:
	 \eqs{
 	\alpha^\chi( \mu ) = \sum_{u=1}^2  \indic \{ \mu \in B_u \} \PP_{\mu}[ S^\chi = u ] .
 }
If $\mu \notin B_1 \cup B_2$, then $\alpha^\chi( \mu ) = 0$ so that the risk is indeed smaller than $\zeta$. Now we assume that $\mu \in B_u$ and we derive an upper bound of $\PP_{\mu}[ S^\chi = u ]$. By definition of IT$_K$, the event $S^\chi = u$ implies that there exists $n \leq s$ such that $\bar{t}^\chi(n)i_u( \hat\mu(n)) \geq f(s,\zeta)$. Using the following two facts: (a) $\mu \in B_u$ and (b) $t_k^\chi(n) \geq \bar{t}^\chi(n)$, we have
	\als{
	f(s,\zeta) &\leq \bar{t}^\chi(n)i_s( \hat\mu(n)) = \bar{t}^\chi(n) \inf_{\lambda \in B_u} \sum_{k=1}^K \KL( \hat\mu_k(n),\lambda(x_k)) \sk
	     &\overset{(a)}{\leq} \bar{t}^\chi(n) \sum_{k=1}^K \KL( \hat\mu_k(n), \mu(x_k)) \overset{(b)}{\leq}  \sum_{k=1}^K t_k^\chi(n) \KL( \hat\mu_k(n) , \mu(x_k)).
	}
	Therefore we have proven that:
	\eqs{
		\alpha^\chi( \mu )  \leq \PP_{\mu} \left[ \sup_{n \leq s} \sum_{k=1}^K t_k^\chi(n) \KL( \hat\mu_k(n) , \mu(x_k))  \geq f(s,\zeta) \right]
	}
	Applying Theorem~\ref{th:kl_concentr} (presented at the end of the appendix) with $\delta := f(s,\zeta)$, we obtain:
	 \eqs{
		\alpha^\chi( \mu ) \leq  e^{K + 1  - f(s,\zeta)} ( f(s,\zeta) \ceil{ f(s,\zeta) \log(s) }/K )^K   \leq \zeta.
		}
		The above inequality holds for all $\mu \in {\cal U}$, and hence the minimax risk satisfies $\alpha^\chi \leq \zeta$, which concludes the proof of (i).

\medskip
\noindent
\underline{Proof of (ii) (Expected length).}
	
We now consider $1 \leq k \leq K$ and we derive an upper bound of $\EE_{\mu}[ t_k^\chi(s)]$. Fix $\epsilon > 0$, and define $t_0 = (1+\epsilon) f(s,\zeta) / i_u(\mu(x_1),\dots,\mu(x_K))$. Introduce the following two sets of rounds:
	\als{
	A &= \{ 1 \leq n \leq s:  x(n) = x_k , \overline{t}^\chi(n) \leq t_0 \}, \sk
	B &= \{ 1 \leq n \leq s:  x(n) = x_k , \overline{t}^\chi(n) \geq t_0 \}.
	}
We have $t_k^\chi(s) \leq |A| + |B|$. Furthermore, in each round $n \in A$, $t_k^\chi(n)$ is incremented, therefore $|A| \leq t_0$. Now let $n \in B$. By design of IT$_K$, this implies that: $t_0 \leq \bar{t}^\chi(n)$ and $\bar{t}^\chi(n)i_u( \hat\mu(n) ) \leq f(s,\zeta)  $. Therefore:
$$
t_0 i_u( \hat\mu(n)) \leq f(s,\zeta),
$$
and thus:
\eq{\label{eq:elim1}
i_u( \hat\mu(n)) \leq i_u( \mu(x_1),\dots,\mu(x_K) ) / (1+\epsilon). 
}
Now one can verify that the function $(\lambda_1,\dots,\lambda_K) \mapsto \sum_{k=1}^K \KL( \mu(x_k),\lambda_k )$ attains its infimum on $B_u$. By continuity of $\KL$ in its second argument, there must exist $\lambda^\star \in B_u$ such that:
	\eqs{
	    i_u( \mu(x_1),\dots,\mu(x_K)) =  \sum_{k=1}^K \KL( \mu(x_k),\lambda^\star(x_k) ).
	}
Let $\eta > 0$ such that we have $|\hat\mu_k(n) - \mu(x_k)  | \leq \eta$ for all $k$. Since $\lambda^\star \in B_u$, this implies that:
	\eq{
		i_u( \hat\mu(n)) = \inf_{\lambda \in B_u}  \sum_{k=1}^K \KL( \hat\mu_k(n),\lambda(x_k) ) \leq \sum_{k=1}^K \KL( \hat\mu_k(n),\lambda^\star(x_k) ). \label{eq:elim2}
	}
Since $|\hat\mu_k(n) - \mu(x_k)  | \leq \eta$ for all $k$, the r.h.s. of \eqref{eq:elim2} tends to \newline $i_u( \mu(x_1),\dots,\mu(x_K) ) < i_u( \mu(x_1),\dots,\mu(x_K) ) / (1+\epsilon)$ as $\eta \to 0$. Hence the inequality~\eqref{eq:elim1} cannot hold for arbitrary small $\eta$. 
	
Hence, there exists $\eta_0$ such that $n \in B$ implies $\max_k | \hat\mu_k(n) - \mu(x_k) | \geq \eta_0$. Note that $\eta_0$ might depend on $\epsilon$ and $\mu(x_1), \dots, \mu(x_K)$. Using Lemma~\ref{lem:loglog}, we get $\EE[|B|] = o( \log(s) )$.
		
Therefore we have:
\eqs{
\EE[t_k^\chi(s)] \leq  \frac{(1 + \epsilon)f(s,\zeta)}{i_u( \mu(x_1),\dots,\mu(x_K) )}   +  o(\log(s)).
}
As noted in remark~\ref{rem:fbar}, when considering $\zeta = s^{-\gamma}$, we may use $f(s,\zeta) = \gamma \log(s) + o(\log(s))$, hence:
\eqs{
\lim\sup_{s \to \infty} \frac{ \EE[t_k^\chi(s)] }{\log(s)} \leq \frac{(1 + \epsilon)\gamma}{i_u( \mu(x_1),\dots,\mu(x_K) )}. 
}
	Since the above inequality holds for all $\epsilon > 0$, we obtain the announced result:
		\eqs{
		\lim\sup_{s \to \infty} \frac{ \EE[t_k^\chi(s)] }{\log(s)} \leq \frac{\gamma}{i_u( \mu(x_1),\dots,\mu(x_K) )}. 
	}
	which concludes the proof of (ii).

\subsection{Proof of Theorem~\ref{th:IT3}}

We start by proving Lemma~\ref{lem:is_lowerbound} which shows that $i_u$ can be lower bounded by the $\text{KL}^{\star}$ function.
\begin{lemma}\label{lem:is_lowerbound}
Consider Bernoulli rewards. Define $m=1$ if $u=1$ and $m=3$ otherwise. Then we have for all $u \in \{1,2\}$:
	\eqs{
	i_u( \mu(x_1),\mu(x_2),\mu(x_3)) \geq \KL^{\star}(\mu(x_m), \mu(x_2)).
	}
\end{lemma}
\bp
	We only prove the statement for $u=1$, as the case $u=2$ follows by symmetry. By a slight abuse of notation we denote $\mu(x_k)$ and $\lambda(x_k)$ by $\mu_k$ and $\lambda_k$ respectively.
	
First note that if $\mu_2 < \mu_1$, we have $\text{KL}^{\star}(\mu_1, \mu_2) = 0$ and the statement holds because \newline $i_u( \mu(x_1),\mu(x_2),\mu(x_3)) \geq 0$, since the KL divergence is positive. 

Now consider the case $\mu_2 \geq \mu_1$. We have the inequality:
\eqs{
i_1( \mu_1,\mu_2,\mu_3) =  \inf_{\lambda \in B_1} \sum_{k=1}^3 \KL( \mu_k,\lambda_k) \geq \inf_{\lambda \in B_1} \sum_{k=1}^2 \KL(\mu_k,\lambda_k).
}
Define function $\phi :[0,1]^2 \to \RR$ by $\phi(	\lambda_1,\lambda_2) = \sum_{k=1}^2 \KL( \mu_k,\lambda_k)$. Define the set $\Lambda = \{ (\lambda_1,\lambda_2): \lambda_1 \geq \lambda_2 \}$. Consider $\lambda \in B_1$, then $x \mapsto \lambda(x)$ attains its maximum in $[\underline{x},x_1]$, and since $\lambda$ is unimodal we must have $\lambda_1 \geq \lambda_2$. Therefore:
\al{
i_1( \mu_1,\mu_2,\mu_3) &\geq  \min_{ (\lambda_1,\lambda_2) \in  \Lambda } \phi(\lambda_1,\lambda_2). \label{eq:elimfinite1}
}
Consider $(\lambda_1^\star,\lambda_2^\star) \in \arg \min_{(\lambda_1,\lambda_2) \in \Lambda } \phi(\lambda_1,\lambda_2)$. We are going to prove that we must have $\lambda_1^\star=\lambda_2^\star$. Consider two subcases (a) $0 \leq \lambda_2^\star \leq \mu_1$ and (b) $\mu_1 \leq \lambda_2^\star \leq 1$. In case (a) we must have $\lambda_1^\star = \mu_1$ since $\lambda_1 \mapsto \KL(  \mu_1 ,\lambda_1) $ attains its minimum at $\mu_1$. In turn we must have $\lambda_2^\star = \mu_1 = \lambda_1^\star$ since $\lambda_1 \mapsto \KL(  \mu_1 ,\lambda_1)$ is decreasing for $\lambda_1 \leq \mu_1 \leq \mu_2$. In case (b), we must have $\lambda_1^\star = \lambda_2^\star$ because $\lambda_1 \mapsto \KL(  \mu_1 ,\lambda_1)$ is increasing for $\lambda_1 \geq \lambda_2^\star \geq \mu_1$.  In both cases we have proven that $\lambda_1^\star = \lambda_2^\star$.

Define function $\tilde{\phi}(\lambda) = \phi(\lambda,\lambda)$, from the reasoning above we have that:
\eqs{
 \min_{ (\lambda_1,\lambda_2) \in \Lambda } \phi(\lambda_1,\lambda_2) = \min_{ \lambda \in [0,1] } \tilde{\phi}(\lambda).
}
\begin{itemize}
\item If $\mu_1=\mu_2=0$, then $\phi(0,0) = 0$ so that the optimum is $\lambda^\star = 0$.
\item If $\mu_1=\mu_2=1$, then $\phi(1,1) = 0$, so that the optimum is $\lambda^\star = 1$.
\item Otherwise, denote by $\tilde{\phi}'$ the first derivative of $\tilde{\phi}$. We have:
\end{itemize}
\eqs{
	\tilde{\phi}'(\lambda) = \frac{2 - (\mu_1 + \mu_2)}{1 - \lambda} - \frac{\mu_1 + \mu_2}{\lambda}.
}
and $\tilde{\phi}'(0^+) = -\infty$ and $\tilde{\phi}'(1^-) = +\infty$  so that $\tilde{\phi}$ attains its maximum in the interior of $[0,1]$. Solving for $\tilde{\phi}'(\lambda^\star) = 0$ we obtain the unique solution $\lambda^\star = (\mu_1 + \mu_2)/2$. 

We observe that in the three above cases, the optimum is $\lambda^\star =(\mu_1 + \mu_2)/2$. We have proven the announced inequality:
\eqs{
i_1( \mu(x_1),\dots,\mu(x_K)) \geq \min_{ (\lambda_1,\lambda_2) \in \Lambda } \phi(\lambda_1,\lambda_2) = \min_{ \lambda \in [0,1] } \tilde{\phi}(\lambda) = \tilde{\phi}( (\mu_1 + \mu_2)/2) = \text{KL}^{\star}(\mu_1, \mu_2)   .
}
\ep

\medskip
\noindent
\textbf{ Proof of Theorem~\ref{th:IT3}. }\\
\underline{(i) Minimax risk of \subalgom$_3$.} The minimax risk of \subalgom$_3$ is upper bounded by $\zeta$ by Theorem~\ref{th:elim}.\\
\underline{(i)' Minimax risk of $\chi=$IT$_3'$.} Let $\mu \in B_u$, let us upper bound $\PP_\mu[S^{\chi} = u]$. Without loss of generality consider $u=1$ and a time instant $n \leq s$ such that $S^{\chi}(n) = 1$. By definition of \subalgoprimem$_3$ this implies that $\bar{t}^{\chi}(n) \text{KL}^{\star}(\hat\mu_1(n),\hat\mu_2(n)) \geq f(s,\zeta)$. We deduce that:
\als{
	f(s,\zeta) &\leq \bar{t}^{\chi}(n) \text{KL}^{\star}(\hat\mu_1(n),\hat\mu_2(n)) \sk
			    &\overset{(a)}{\leq} \bar{t}^{\chi}(n) i_1( \hat\mu_1(n),\dots, \hat\mu_K(n)) \sk
                &= \bar{t}^\chi(n) \inf_{\lambda \in B_1} \sum_{k=1}^K \KL( \hat\mu_k(n),\lambda(x_k)) \sk
	           &\overset{(b)}{\leq} \bar{t}^\chi(n) \sum_{k=1}^K \KL( \hat\mu_k(n), \mu(x_k)) \sk
	           &\overset{(c)}{\leq} \sum_{k=1}^K t_k^\chi(n) \KL( \hat\mu_k(n) , \mu(x_k)).
	}
where we have used (a) Lemma~\ref{lem:is_lowerbound}, (b) the fact that $\mu \in B_1$ (c) the fact that $\bar{t}^\chi(n) \leq t_k^\chi(n)$ for all $k$. Applying theorem~\ref{th:kl_concentr} once again:
	\eqs{
		\alpha^\chi( \mu )  \leq \PP \left[ \sup_{n \leq s} \sum_{k=1}^K t_k^\chi(n) \KL( \hat\mu_k(n) , \mu(x_k))  \geq f(s,\zeta) \right] \leq \zeta
	}
which proves that $\alpha^\chi( \mu ) \leq \zeta$ for all $\mu \in {\cal U}$ and concludes the proof of (i)'.

\medskip
\noindent
\underline{(ii) Expected duration of $\chi=$IT$_3$.} The proof of (ii) for IT$_3'$ follows by the same arguments. By a slight abuse of notation we denote $\mu(x_k)$ by $\mu_k$. Without loss of generality, consider $\mu$ such that $x^\star \in [x_2,\overline{x}]$. Therefore we have that $\mu_2 > \mu_1$ since $\mu$ is unimodal.  Fix $0 < \epsilon < \delta/2$, and define $t_0 = f(s,\zeta) / \text{KL}^{\star,\epsilon}(\mu_1,\mu_2)$. Introduce the two sets of instants:
	\als{
	A = \{ 1 \leq n \leq s:  x(n) = x_k , \bar{t}^\chi(n) \leq t_0 \} \;\;,\;\; B = \{ n \geq 1:  x(n) = x_k , \max_{k' \in \{1,2\} } | \hat\mu_{k'}(n) - \mu_{k'} | \geq \epsilon \}.
	}
We prove that $x(n) = x_k$ implies that $n \in A \cup B$. Consider $n$ such that $\bar{t}^\chi(n) \geq t_0$ and $|\hat\mu_{k'}(n) - \mu_{k'}| \leq \epsilon$ , $k'\in \{1,2\}$. Since $\epsilon < \delta/2 \leq (\mu_2-\mu_1)/4$ we have: 
	\als{
		 \hat\mu_1(n) &\leq \mu_1 + \epsilon \leq (\mu_1 + \mu_2)/2 - \epsilon \leq (\hat\mu_1(n) + \hat\mu_2(n))/2 \sk
		 \hat\mu_2(n) &\geq \mu_2 - \epsilon \geq (\mu_1 + \mu_2)/2 + \epsilon \geq (\hat\mu_1(n) + \hat\mu_2(n))/2
	}
so that $\KL^{\star}( \hat\mu_1(n),\hat\mu_2(n)) \geq \KL^{\star,\epsilon}( \mu_1,\mu_2)$. 
Applying Lemma~\ref{lem:is_lowerbound}, we have:
	\als{
		\bar{t}(n) i_1(\hat\mu(n)) \geq \bar{t}^\chi(n) \KL^{\star}( \hat\mu_1(n),\hat\mu_2(n)) \geq  t_0 \KL^{\star,\epsilon}( \mu_1,\mu_2) = \overline{f}(s).
	}
Therefore we cannot have $x(n) = x_k$.
	
	We have proven that $t_k^\chi(s) \leq |A| + |B|$. Furthermore, at each instant $n \in A$, $\bar{t}^\chi(n)$ is incremented, therefore $|A| \leq t_0$. Let us upper bound the expected size of $B$. Decompose $B = B^1 \cup B^2$, with:
	\eqs{
	B^{k'} = \{ n \geq 1:  x(n) = x_k ,  | \hat\mu_{k'}(n) - \mu_{k'} | \geq \epsilon \}.
	}
	Let $n \in B^{k'}$ and define $a = \sum_{n' \leq n} \indic\{ n' \in B^{k'} \}$ so that $n$ is the $a$-th instant of $B^{k'}$. Then we have that $t_{k'}^\chi(n) \geq a$ and applying \cite{ICMLtechreport}[Lemma 2.2] we have that for $k' \in \{1,2\}$, $\EE[|B^{k'}|] \leq \epsilon^{-2}$. Therefore $\EE[|B|] \leq 2 \epsilon^{-2}$. So statement (ii) is proven:
	\eqs{
		\EE_{\mu}[ t_k^{\chi}(s) ] \leq  t_0 + 2 \epsilon^{-2} = \frac{\overline{f}(s)}{\text{KL}^{\star,\epsilon}( \mu(x_1),\mu(x_2)) } + 2 \epsilon^{-2}.
	}
	
\medskip
\noindent
\underline{(iii) Further bounds on the duration of $\chi=$IT$_3$.} The proof of (iii) for IT$_3'$ follows by the same arguments. To establish the announced inequalities, we will use the following fact: from Pinsker's inequality $\KL(\alpha,\beta) \geq 2(\alpha-\beta)^2$ for all $(\alpha,\beta) \in [0,1]^2$, so that:
	\eqs{
	\KL^{\star,\epsilon}(\mu_1,\mu_2 ) \geq 4 ( (\mu_2-\mu_1)/2 - 2 \epsilon)^2 \geq 4(\delta - 2 \epsilon)^2, 
	}
	In particular for $\epsilon = \delta/4$ we have $\KL^{\star,\epsilon}(\mu_1,\mu_2 ) \geq \delta^2$.
	
\noindent	
\underline{Inequality (a).} 
	Define $t_0 = 8 \overline{f}(s) \delta^{-2}$ and $n_0 = 3 t_0$. By design of IT$_3$, for all $k$ we have $t_k^\chi(n_0) = t_0$. Set $\epsilon = \delta/4$. If both $\hat\mu_1(n_0) \leq \mu_1 + \epsilon$ and $\hat\mu_2(n_0) \geq \mu_2 - \epsilon$ then we have:
	\eqs{
	\bar{t}^\chi(n_0) \KL^{\star}(\hat\mu_1(n_0),\hat\mu_2(n_0)) \geq t_0 \KL^{\star,\epsilon}(\mu_1,\mu_2 ) \geq 8 \overline{f}(s) \delta^{-2} \delta^2 = 8 \overline{f}(s) > \overline{f}(s).
	}
so that IT$_3$ must terminate at time $n_0$ or before. Hence, applying Hoeffding's inequality:
\eqs{
\PP_{\mu}[  t_k^\chi(s) \geq t_0] \leq \PP[ \hat\mu_k(n_0) \geq \mu_1 + \epsilon  ] + \PP[\hat\mu_2(n_0) \leq \mu_2 - \epsilon] 													\leq 2e^{-2 t_0 \epsilon^2} = 2 e^{-\overline{f}(s)}.
}
which is the announced result.\\
\noindent
\underline{Inequality (b).} Once again setting $\epsilon = \delta/4$, and using both $\KL^{\star,\epsilon}(\mu_2,\mu_1 ) \geq \delta^2$ and statement (ii), we obtain the second claim:
	\eqs{
	\EE_{\mu}[ t_k^{\chi}(s) ] \leq \frac{\overline{f}(s) + 32}{\delta^2} 
	}
\\
\noindent
\underline{Inequality (c).} By statement (ii), and using the fact that $\overline{f}(s) = \gamma \log(s) + o(\log(s))$, for all $\epsilon > 0$, we have:
	\eqs{
	\lim\sup_{s \to \infty}	\frac{ \EE_{\mu}[ t_k^{\chi}(s) ]}{\log(s) } \leq \frac{\gamma}{\text{KL}^{\star,\epsilon}( \mu(x_1),\mu(x_2)) },
	}
	so that letting $\epsilon \to 0$ in the above expression yields:
	\eqs{
	\lim\sup_{s \to \infty}	\frac{ \EE_{\mu}[ t_k^{\chi}(s) ]}{\log(s) } \leq \frac{\gamma}{\text{KL}^{\star}( \mu(x_1),\mu(x_2)) },
	}
	which concludes the proof of statement (iii).
\ep

\subsection{Proof of Theorem~\ref{th:regret}}

Fix $N$ throughout the proof. We introduce the following notations. The algorithm proceeds in phases, each phase corresponding to a call of IT$_3$ (or IT$_3'$) subroutine. We define $I^{N'}$ the interval output after the $N'$-th call of IT$_3$, with $I^{0} = [0,1]$. We define $\tau^{N'}$ the duration of the $N'$-th call of IT$_3$. Define the event:
	\eqs{
	A = \cap_{N'=0}^{N} \{ x^\star \in I^{N'} \},
	}
which corresponds to sample paths where the first $N$-th calls of IT$_3$ have returned an interval containing the optimal arm $x^\star$. We denote by $A^c$ the complement of $A$.

The regret due to sample paths in $A^c$ is upper bounded by $\mu^\star T \PP[ A^c ]$. The regret due to the $N'$-th phase for sample paths in $A$ is upper bounded by $\EE[ \tau^{N'} \indic\{ A \} ( \mu^\star - \min_{x \in I^{N'}} \mu(x))]$. This is true because the $N'$-th phase has duration $\tau^{N'}$, and during that phase only arms in $I^{N'}$ are sampled so that the regret of a sample in $I^{N'}$ is upper bounded by $\mu^\star - \min_{x \in I^{N'}} \mu(x)$. Therefore the regret admits the following upper bound:
\eqs{
R^{\pi}(T) \leq \mu^\star T \PP[A^c] + \sum_{N' \geq 0} \EE[ \tau^{N'} \indic\{ A \} ( \mu^\star - \min_{x \in I^{N'}} \mu(x) ) ].
}
Consider a sample path in $A$, and $N' \leq N$, then we have $|I^{N'}| \leq \psi^{N'}$ and $x^\star \in I^{N'}$. Therefore $\mu^\star - \min_{x \in I^{N'}} \mu(x) \leq g_\mu( \psi^{N'} )$ by definition of $g_{\mu}$. Similarly, consider a sample path in $A$, and $N' > N$. Then we have $I^{N'} \subset I^{N}$, $|I^{N}| \leq \psi^{N}$ and $x^\star \in I^{N}$. Therefore:
\eqs{
\mu^\star - \min_{x \in I^{N'}} \mu(x)  \leq \mu^\star - \min_{x \in I^{N}} \mu(x) \leq g_\mu( \psi^N ),
}
and the regret satisfies:
\als{
R^{\pi}(T) &\leq \mu^\star T \PP[A^c] +  \sum_{N' = 0}^{N} g_\mu( \psi^{N'} ) \EE[ \tau^{N'} \indic\{ A\}]  + g_\mu( \psi^{N} ) \sum_{N' > N} \EE[ \tau^{N'} \indic\{ A \}]  \sk
&\leq \mu^\star T \PP[A^c]  + \sum_{N' = 0}^{N} g_\mu( \psi^{N'} ) \EE[ \tau^{N'} \indic\{ A \}] + g_\mu( \psi^{N} ) \EE[ \sum_{N' > N} \tau^{N'} ], \sk
&\leq \mu^\star T \PP[A^c]  + \sum_{N' = 0}^{N} g_\mu( \psi^{N'} ) \EE[ \tau^{N'} \indic\{ A \}] + T g_\mu( \psi^{N} ),
}
where we have used the fact that $\sum_{N' > N} \tau^{N'} \leq \sum_{N' \geq 0} \tau^{N'} = T$. 

We now upper bound the probability of event $A^c$. Since $x^\star \in I^{0} = [0,1]$, the occurrence of $A^c$ implies that there exists $N' < N$ such that $x^\star \in I^{N'}$ and $x^\star \not\in I^{N'+1}$ so that we have the inclusion:
\eqs{
A^c \subset \cup_{N'=0}^{N-1}  \{ x^\star \in I^{N'} , x^\star \notin I^{N'+1}   \}.
}
Since the event $\{ x^\star \in I^{N'} , x^\star \notin I^{N'+1}   \}$ corresponds to an incorrect decision taken under IT$_3$, we have $\PP[  x^\star \in I^{N'} , x^\star \notin I^{N'+1}   ] \leq T^{-\gamma}$, because of Theorem~\ref{th:IT3}. Using a union bound we obtain the upper bound:
\eqs{
\PP[A^c] \leq \sum_{N'=0}^{N-1} \PP[  x^\star \in I^{N'} , x^\star \notin I^{N'+1}   ] \leq N T^{-\gamma}.
}
The regret upper bound becomes:
\eqs{
R^{\pi}(T) \leq  \mu^\star N T^{1-\gamma}  + T g_\mu( \psi^{N} ) + \sum_{N' = 0}^{N} g_\mu( \psi^{N'} ) \EE[ \tau^{N'} \indic\{ A \}] .
}
Finally, from Theorem~\ref{th:IT3}, we have that $\EE[ \tau^{N'} \indic\{ A \}] \leq 3 ( \overline{f}(T) + 32)(\delta(I^{N'}))^{-2}$ (we sample from $3$ arms) where $\delta(I^{N'})$ is the quantity $\delta$ defined in the statement of Theorem 3, when the interval considered by IT$_3$ is $I^{N'}$. Since we are considering a sample path in $A$, and $N' \leq N$ we have once again that $|I^{N'}| \leq \psi^{N'}$ and $x^\star \in I^{N'}$ so that $\delta(I^{N'}) \geq h_{\mu}(\psi^{N'})$ by definition of $h_\mu$. Therefore: $ \EE[ \tau^{N'} \indic\{ A \}] \leq 3 (\overline{f}(T) + 32)( h_{\mu}(\psi^{N'}) )^{-2}$. We obtain finally: 
\eqs{
R^{\pi}(T) \leq  \mu^\star N T^{1-\gamma}   + T g_\mu( \psi^{N} ) + 3 (\overline{f}(T) + 32) \sum_{N' = 0}^{N} g_\mu( \psi^{N'} ) ( h_{\mu}(\psi^{N'}) )^{-2},
}
which is the announced result and concludes the proof.

\subsection{Proof of Theorem~\ref{th:regretclass}}
	To prove Theorem~\ref{th:regretclass}, we use the following intermediate result.
	\begin{proposition}\label{prop:1}
		For all $\mu \in {\cal U}(C_1,C_2)$:
		
		(a) $g_{\mu}(\Delta) \leq C_{2} \Delta^{\xi}$;
		
		(b) $h_{\mu}(\Delta) \geq C_{1} a_{\xi} \Delta^{\xi}$, with $a_{\xi} = 4^{-\xi} \min( 1 , 2^{\xi} - 1)$
	\end{proposition}
	\bp
	(a) By definition of $g_\mu$ and since $\mu \in {\cal U}(C_1,C_2)$, we have: 
	\eqs{
	g_\mu(\Delta) = \mu^\star - \min( \mu( x^\star - \Delta ) , \mu( x^\star + \Delta ) ) \leq C_{2} \Delta^{\xi}.
	}
	
	(b) Consider $x$ such that $x^\star \leq x \leq x^\star + \Delta/4$. Since since $\mu \in {\cal U}(C_1,C_2)$, we have:
	\eqs{
		\mu(x) - \mu(x + \Delta/4) \geq C_{1} ( ( x + \Delta/4 - x^\star)^{\xi} - (x - x^\star)^{\xi}).
	} 
Fix $\Delta$, and define the function $l(x) = (x + \Delta/4 - x^\star)^{\xi} - (x - x^\star)^{\xi}$. Its first derivative is:
	\eqs{
		l'(x) = \xi (  (x + \Delta/4 - x^\star)^{\xi-1} -  (x - x^\star)^{\xi-1} ).
	}
	Therefore the function $x \mapsto l(x)$ on interval $[x^\star , x^\star + \Delta/4]$ is increasing if $\xi \geq 1$ and decreasing if $\xi < 1$ so we get the lower bound:
	\eqs{
		\min_{x \in [x^\star , x^\star + \Delta/4]} \mu(x) - \mu(x + \Delta/4) \geq \begin{cases} C_{1} l(x^\star) = C_{1} (\Delta/4)^{\xi}   & \text{ if  } \xi \geq 1 \\  C_{1} l(x^\star + \Delta/4) = C_{1} (2^{\xi} - 1) (\Delta/4)^{\xi} & \text{ if  }  \xi < 1 \end{cases}
	}
	so that $h_{\mu}(\Delta) \geq C_{1} \min( 1 , 2^{\xi} - 1) (\Delta/4)^{\xi}$ as announced.
\ep
	
Let us now prove Theorem~\ref{th:regretclass}. From Theorem~\ref{th:regret}, we can decompose the regret upper bound into three terms:
\als{
R^{\pi}(T) &\leq  r_1(T) + r_2(T) + r_3(T) \sk
	r_1(T) &=\mu^\star N T^{1-\gamma} \sk
	r_2(T) &= T g_\mu( \psi^{N} ) \sk
	r_3(T) &= 3 (\gamma \log(T) + 32) \sum_{N' = 0}^{N} g_\mu( \psi^{N'} ) ( h_{\mu}(\psi^{N'}) )^{-2}.
}
We proceed to upper bound each term. The first term $r_1(T)$ is explicit. By Proposition~\ref{prop:1}, the second term is upper bounded as: $r_2(T) \leq T C_2 \psi^{\xi N}$. As for the third term $r_3(T)$, by Proposition~\ref{prop:1}, we have that $g_\mu( \psi^{N'} ) \leq C_2 \psi^{\xi N'}$ and $h_{\mu}(\psi^{N'}) \geq C_1 a_{\xi} \psi^{\xi N'}$, so that:
\als{
	\sum_{N' = 0}^{N} g_\mu( \psi^{N'} ) ( h_{\mu}(\psi^{N'}) )^{-2} 
	&\leq \sum_{N' = 0}^{N} C_2 (C_1 a_{\xi})^{-2} \psi^{-\xi N'} 
	\leq \frac{C_2 \psi^{-\xi (N+1)}}{C_1^2 a_{\xi}^{2} (\psi^{-\xi} - 1)}.
}
Finally, we get:
\eqs{
R^{\pi}(T) \leq \mu^\star N T^{1-\gamma} + T C_2 \psi^{\xi N} + \frac{3 (\overline{f}(T) + 32) C_2 \psi^{-\xi (N+1)}}{C_1^2 a_{\xi}^{2} (\psi^{-\xi} - 1)}
}
Define $M \geq 0$ (not necessarily an integer) such that the last two terms in the r.h.s. of the above inequality are equal:
\eqs{
	C_2 T \psi^{\xi M} = \frac{3 (\overline{f}(T) + 32) C_2 \psi^{-\xi (M+1)}}{C_1^2 a_{\xi}^{2} (\psi^{-\xi} - 1)}.
}
We have that:
\eqs{
	\psi^{- 2 \xi M} = \frac{T C_1^2 a_{\xi}^2 ( \psi^{-\xi} - 1)}{3 (\overline{f}(T) + 32) \psi^{-\xi} } \leq T C_1^2
}
since $ a_{\xi} \leq 1$, $ \psi^{-\xi} - 1 \leq \psi^{-\xi}$ and $\overline{f}(T) \geq 1$. Taking logarithms we deduce that:
\eqs{
	M \leq \frac{\log( T C_1 )}{\xi \log( 1/\psi )}.
}
Now set $N \equiv \ceil{M}$ for the remainder of the proof. We obtain the announced upper bound:
\als{
	R^{\pi}(T) &\leq \mu^\star (M+1) T^{1-\gamma} + T C_2 \psi^{\xi M} + \psi^{-\xi} \frac{3 (\overline{f}(T) + 32) C_2 \psi^{-\xi (M+1)}}{C_1^2 a_{\xi}^{2} (\psi^{-\xi} - 1)} \sk
				&\leq \mu^\star T^{1-\gamma} \left(\frac{\log( T C_1 )}{\xi \log( 1/\psi )}+1\right) + (1 + \psi^{-\xi}) T C_2 \psi^{\xi M} \sk
				&\leq \mu^\star T^{1-\gamma} \frac{\log( T C_1 \psi^{-\xi})}{\xi \log( 1/\psi )}  +  \frac{2 \psi^{-3\xi/2} C_2}{C_1 a_{\xi}} \sqrt{ \frac{3 T (\overline{f}(T) + 32) }{\psi^{-\xi} - 1}}. 
}
This concludes the proof.

\subsection{Proof of Theorem \ref{th:optimerror}}

The proof proceeds along the same lines as the proof of Theorem~\ref{th:regretclass}. Define $M \geq 0$ such that:
\eqs{
	\frac{24 \overline{f}(T) \psi^{-2 \xi (M+1)}}{ a_{\xi}^2 C_1^{2} (\psi^{-2 \xi} - 1)} = T.
}
Let us first upper bound $M$. We have that:
\eqs{
	\psi^{- 2 \xi(M+1)} = \frac{C_1^2 a_{\xi}^2 T(\psi^{-2\xi}-1)}{24 \overline{f}(T)} \leq C_1^2 T \psi^{-2\xi}.
}
using the fact that $a_{\xi} \leq 1$, $\overline{f}(T) \geq 1$ and $\psi^{-2\xi}-1 \leq \psi^{-2\xi}$. Hence, taking logarithms:
\eqs{
M \leq M + 1 \leq \frac{\log(T C_1 \psi^{-\xi})}{ \xi \log(1/\psi)}.
}
We now fix $N = \floor{M}$ for the remainder of the proof. Once again the algorithm proceeds in phases, each phase corresponding to a call to IT$_3$ (or IT$_3'$). We define $I^{N'}$ the interval output by the $N'$-th call of IT$_3$, with $I^{0} = [0,1]$. We define $\tau^{N'}$ the duration of the $N'$-th call of IT$_3$. We define two events:
	\als{
	A &= \cap_{N'=0}^{N} \{ x^\star \in I^{N'} \}, \sk
	B &= \cap_{N'=0}^{N} \{ \tau^{N'} \leq 24 \overline{f}(T) h_{\mu}(\psi^{N'})^{-2}  \}
	}
	$A$ corresponds to sample paths where the first $N$-th calls of IT$_3$ have returned an interval containing the optimal arm $x^\star$. $B$ corresponds to sample paths where the first $N$-th calls to IT$_3$ have not lasted more than their ``typical length'' (as prescribed by Theorem~\ref{th:IT3}). The optimization error can hence be decomposed according to the occurrence of $A$ and $B$:
	\als{
		E^{\pi}(T) &= \EE[ (\mu^\star - \mu(x(T)) \indic\{ A \cap B \} ] + \EE[ (\mu^\star - \mu(x(T)) \indic\{ (A \cap B)^c \}] \sk
					&\leq \EE[ (\mu^\star - \mu(x(T)) \indic\{ A \cap B \} ] + \mu^\star \EE[ \indic\{ (A \cap B)^c \}  ] \sk
					&\leq  \EE[ (\mu^\star - \mu(x(T)) \indic\{ A \cap B \} ] + \mu^\star(\PP[A^c] + \PP[B^c \cap A]).
	}
	We will establish two facts:
	\begin{itemize}
	\item[(a)] $\PP[A^c] + \PP[B^c \cap A] \leq 3 M T^{-\gamma}$
	\item[(b)] $(\mu^\star - \mu(x(T)) \indic\{ A \cap B \} \leq  C_2 \psi^{\xi (M+1)}$ a.s.
	\end{itemize}
	If (a) and (b) hold we have that:
	\eqs{
		E^{\pi}(T) \leq C_2 \psi^{\xi (M+1)} + 3 \mu^\star M T^{-\gamma}
					\leq  \frac{C_2}{C_1 a_{\xi}} \sqrt{ \frac{24 \overline{f}(T)}{T(\psi^{-2\xi}-1)}} +  \frac{3 T^{-\gamma} \mu^\star \log(T C_1 \psi^{-\xi})}{ \xi \log(1/\psi)}.
	}
which is precisely the announced result.

\fbox{Fact (a)} From Theorem~\ref{th:IT3}, statement (i), we know that $\PP[A^c] \leq N T^{-\gamma}$ since the risk of  IT$_3$ is upper bounded by $T^{-\gamma}$. Furthermore, from Theorem~\ref{th:IT3}, statement (iii a), we know that $\PP[B^c \cap A] \leq 2 N T^{-\gamma}$ since test IT$_3$ applied to an interval of size $\psi^{N'}$ that contains the optimal arm has length greater than $24 \overline{f}(T) h_{\mu}(\psi^{N'})^{-2}$ with probability less than $2e^{-\overline{f}(T)} \leq 2 T^{-\gamma}$. Hence $\PP[A^c] + \PP[B^c \cap A] \leq 3 N T^{-\gamma} \leq 3 M T^{-\gamma}$ as announced.

\fbox{Fact (b)} 	Let us prove that if $B$ occurs, then the first $N$-th calls to IT$_3$ terminate before the time horizon $T$. Indeed, if $B$ occurs, applying Proposition~\ref{prop:1}, one has:
	\als{
		\sum_{N'=0}^{N} \tau^{N'} &\leq 24 \overline{f}(T) \sum_{N'=0}^{N} h_{\mu}(\psi^{N'})^{-2} 
		\leq  \frac{24 \overline{f}(T)}{a_{\xi}^{2} C_1^{2}} \sum_{N'=0}^{N}  \psi^{-2 \xi N'} \sk
		&\leq  \frac{24 \overline{f}(T) \psi^{-2 \xi (N+1)}}{ a_{\xi}^2 C_1^{2} (\psi^{-2 \xi} - 1)}
		\leq \frac{24 \overline{f}(T) \psi^{-2 \xi (M+1)}}{ a_{\xi}^2 C_1^{2} (\psi^{-2 \xi} - 1)}
		= T
	}
	so that the first $N$ tests do terminate before $T$. Furthermore, if $A$ occurs, the $N$-th test returns an arm $x$ such that $|x - x^\star| \leq \psi^{M+1}$. In turn, by proposition~\ref{prop:1}, one has $|\mu^\star - \mu(x)| \leq g_{\mu}(\psi^{M+1}) \leq C_2 \psi^{\xi(M+1)}$. Hence we have proven that, if both $A$ and $B$ occur one has $(\mu^\star - \mu(x(T)) \leq  C_2 \psi^{\xi (M+1)}$, so that $(\mu^\star - \mu(x(T)) \indic\{ A \cap B \} \leq  C_2 \psi^{\xi (M+1)}$ a.s. as announced. This concludes the proof.

\subsection{Proof of Theorem \ref{th:sequential}}
We work with a given sequential test $\chi$ throughout the proof and we omit the superscript ${}^\chi$ for clarity. Without loss of generality, let $u=1$. We work with a fixed parameter $\lambda \in B_1$. We denote by $Y(s) = ( X_1(x(1)), \dots, X_s(x(s)) )$ the observed rewards from round $1$ to round $s$. We denote by $P_s$ and $Q_s$ the probability distribution of $Y(s)$ under $\mu$ and $\lambda$ respectively.
From Lemma~\ref{lem:sum_kls} (stated and proved at the end of the appendix), we have:
\eq{\label{eq:seq1}
\KL( P_s ||  Q_s ) = \sum_{k=1}^K \EE[ t_k(s) ] \KL(\mu(x_k),\lambda(x_k)).	
}
Consider the event $S = 1$. Since the sequential test $\chi$ has minimax risk smaller than $\alpha$, and $\lambda \in B_1$, we have $\PP_{\lambda}[ S = 1 ] \leq \alpha$. Recall that by assumption $\PP_{\mu}[ S = 1 ] = \beta$ and $\alpha \leq \beta$. Now $S$ is a function of $Y(s)$. Using Lemma~\ref{lem:kl_lb} (stated at the end of the appendix):
\eq{\label{eq:seq2}
\KL( P_s ||  Q_s ) \geq \KLtwo( \PP_{\mu}[ S(s) = 1 ] , \PP_{\lambda}[ S(s) = 1 ] ) \geq \KLtwo(\beta,\alpha).
}
where we have used the fact that $\alpha \mapsto \KLtwo(\beta,\alpha)$ is decreasing for $\alpha \leq \beta$. Putting \eqref{eq:seq1} and \eqref{eq:seq2} together, we obtain:
\eqs{
\sum_{k=1}^K \EE[ t_k(s) ] \KL(\mu(x_k),\lambda(x_k)) \geq \KLtwo(\beta,\alpha).
}
Taking the infimum over $\lambda \in B_1$, we obtain the claimed result:
\eqs{
\inf_{\lambda \in B_1} \sum_{k=1}^K \EE[ t_k(s) ] \KL(\mu(x_k),\lambda(x_k)) \geq \KLtwo(\beta,\alpha).
}

\subsection{Proof of Corollary \ref{cor:sequential1}}

Let us denote $\beta_s = \PP_{\mu}[S_s = 1]$, where $S_s$ is the final decision taken under test $\chi^s$. Since $\beta_s \to_{s \to \infty} \beta > 0$ there exists $s_0$ such that for all $s \geq s_0$ we have $\beta_s \geq s^{-\gamma}$. Since $\chi_s$ has minimax risk $\alpha = s^{-\gamma}$, for all $s \geq s_0$, applying Theorem \ref{th:sequential}, we obtain:
	\eq{\label{cor1:eq1}
	\inf_{\lambda \in B_1} \sum_{k=1}^K \EE[ t_k(s) ] \KL(\mu(x_k),\lambda(x_k)) \geq \KLtwo(\beta_s,\alpha) = \KLtwo(\beta_s,s^{-\gamma}).
	}    
  Now by definition of $\KLtwo$, we have that:
   \eqs{
 	\KLtwo(\beta_s,s^{-\gamma}) = \beta_s \log(\beta_s) + \beta_s \gamma \log(s) + (1 - \beta_s) \log(1 - \beta_s) + (1 - \beta_s) \log(1 - s^{-\gamma}).
 	}
 Since $\beta_s \to_{s \to \infty} \beta > 0$, we have that $ \KLtwo(\beta_s,s^{-\gamma}) \sim_{s \to \infty} \gamma \beta \log(s) $. Letting $s \to \infty$ in~\eqref{cor1:eq1} we have:
 	\eqs{
	\lim \inf_{s \to \infty} \inf_{\lambda \in B_1} \sum_{k=1}^K \frac{\EE_{\mu}[ t_k(s) ]}{\log(s)}  \KL(\mu(x_k),\lambda(x_k)) \geq \gamma \beta,
	}
 which concludes the proof.
 
\subsection{Proof of Corollary \ref{cor:sequential2}}

The proof is constructive: we exhibit a function $\mu$ such that $\PP_{\mu}[S^\chi \neq 0] \geq 1/2$. Without loss of generality we consider interval $I =  [0,1]$. Consider the function $\mu(x) = 1 - 2|1/2 - x|$. $\mu$ is clearly unimodal, with $x^\star = 1/2$ and $\mu^\star = 1$. 

We proceed by contradiction. Consider a test $\chi$ such that $\PP_{\mu}[S^\chi \neq 0] \geq 1/2$. Since $S^\chi \in \{0,1,2\}$, there exists $u \in \{1,2\}$ such that $\PP_{\mu}[S^\chi = u] \geq 1/4$. Without loss of generality consider $u=1$. Let $\epsilon > 0$, and define the function $\lambda^\epsilon$ which is linear on intervals $ \{ [x_1,x_2] , [x_2,x_3], [x_2,(x_3+x_4)/2] ,[(x_3+x_4)/2,x_4]$ with $\lambda(x_k) = \mu(x_k)$, $k \neq 3$ and $\lambda(x_3) = \mu(x_2) + \epsilon$, and $ \lambda((x_3 + x_4)/2) = 1$. One can check that $\lambda^\epsilon$ is unimodal, and attains its maximum in $[x_3,x_4]$. We recall that $\alpha < 1/4$ and applying Theorem \ref{th:sequential}, we obtain the following inequality:
\eqs{
	\sum_{k=1}^K \EE_{\mu}[ t_k(s) ]  \KL(\mu(x_k),\lambda(x_k)) \geq \KLtwo(1/4,\alpha).
}
Since $\KL(\mu(x_k),\lambda(x_k)) = \KL(\mu(x_k),\mu(x_k))= 0$, for $k \neq 3$, and $t_3(s) \leq s$ we obtain:
\eq{\label{cor2:eq1}
	s \KL(\mu(x_3),\mu(x_3) + \epsilon) \geq \KLtwo(1/4,\alpha).
}
Since $\alpha < 1/4$ we have that $\KLtwo(1/4,\alpha) > 0$. On the other hand $\epsilon \mapsto \KL(\mu(x_3),\mu(x_3) + \epsilon)$ is continuous, and $\KL(\mu(x_3),\mu(x_3)) = 0$. Therefore inequality~\eqref{cor2:eq1} cannot hold for all $\epsilon > 0$. This is a contradiction and proves that a test $\chi$ as considered here cannot exist, which concludes the proof.

\subsection{Technical results}

Lemma~\ref{lem:kl_lb} gives a lower bound of the KL divergence of probability measures using the KL divergence between two Bernoulli distributions.
\begin{lemma}\label{lem:kl_lb}
Let $P$ and $Q$ be two probability measures on a probability space $(\Omega,{\cal F}, \PP )$. Assume that $P$ and $Q$ are both absolutely continuous with respect to measure $m(dx)$.
Then:
\eqs{
\KL( P || Q ) \geq \sup_{A \in {\cal F}} \KLtwo(P(A), Q(A)).
}
\end{lemma}
\bp
The proof is based on the log-sum inequality. We recall the derivation of the log-sum inequality here. Consider $f(x) = x \log(x)$. We have that $f''(x) = 1/x$, so that $f$ is convex. We define $p,q$ the densities of $P,Q$ with respect to measure $m$. Then for all $A \in {\cal F}$:
\als{
\int_{A} \log \Lp \frac{ p(x) }{q(x)} \Rp p(x)m(dx)  &=  \int_{A}  f \Lp \frac{ p(x) }{q(x)} \Rp  q(x) m(dx) \sk 
 &=  Q(A) \int_{A} f \Lp \frac{ p(x) }{q(x)} \Rp \frac{ q(x) }{Q(A)} m(dx) \sk
	& \overset{(a)}{\geq} Q(A) f \Lp \int_{A} \frac{ p(x) }{q(x)}  \frac{q(x)}{Q(A)} m(dx) \Rp \sk
	&= Q(A) f \Lp \frac{P(A)}{ Q(A)} \Rp = P(A) \log \Lp \frac{P(A)}{ Q(A)} \Rp. 
}
and (a) holds because of Jensen's inequality.
Applying the reasoning above to $A$ and $A^c = \Omega \setminus A$: 
\als{
\KL(P||Q) &=  \int_{\Omega} \log \Lp  \frac{ p(x) }{q(x)} \Rp  p(x) m(dx) \sk
 &= \int_{A} \log \Lp   \frac{ p(x) }{q(x)} \Rp p(x)m(dx) + \int_{A^c} \log \Lp  \frac{ p(x) }{q(x)}  \Rp p(x)m(dx) \sk
 &\geq P(A) \log \Lp \frac{P(A)}{ Q(A)} \Rp + P(A^c) \log \Lp \frac{P(A^c)}{ Q(A^c)}   \Rp \sk
 &= P(A) \log \Lp \frac{P(A)}{ Q(A)} \Rp + (1 - P(A)) \log \Lp \frac{1-P(A)}{ 1-Q(A)}  \Rp \sk
 &= \KLtwo(P(A),Q(A)).
}
So for all $A$ we have:
\eqs{
\KL(P||Q) \geq \KLtwo(P(A),Q(A)),
}
and taking the supremum over $A \in {\cal F}$ concludes the proof.
\ep

Lemma~\ref{lem:sum_kls} evaluates the KL divergence between sample paths of a given test under two different parameters. The proof follows from a straightforward conditioning argument and is omitted here.
\begin{lemma}\label{lem:sum_kls}
We denote by $Y(s) = ( X_1(x(1)), \dots, X_s(x(s)) )$ the observed rewards from time $1$ to $s$. Consider $\mu,\lambda \in {\cal U}$, and  denote by $P_s$ and $Q_s$ the probability distribution of $Y(s)$ under $\mu$ and $\lambda$ respectively. Then we have:
	\eqs{
	\KL( P_s || Q_s ) = \sum_{k=1}^K \EE[ t_k(s) ] \KL(\mu(x_k),\lambda(x_k)).
	}
\end{lemma}

Theorem~\ref{th:kl_concentr} is a concentration inequality for sums of KL divergences. It was derived derived in~\cite{magureanu2014}, and is stated here for completeness.
\begin{theorem}\label{th:kl_concentr} \cite{magureanu2014}
For all $\delta \geq (K+1)$ and $s \geq 1$ we have:
\begin{equation}\label{eq:con}
\PP\left[ \sup_{n \leq s} \sum_{k=1}^K  t_k(n) \KL (\hat\mu_k(n) , \mu(x_k)  ) \geq \delta \right] \leq e^{K+1 -\delta} \left( \frac{\ceil{\delta \log(s)} \delta}{K}\right)^K.
\end{equation}
\end{theorem}

Lemma~\ref{lem:loglog} is a technical result showing that the expected number of times the empirical mean of i.i.d. variables deviates by more than $\delta$ from its expectation is $o(\log(n))$, $n$ being the time horizon.
\begin{lemma}\label{lem:loglog}
Let $\{X_n\}_{n \geq 1}$ be a family of i.i.d. random variables with common expectation $\mu$ and finite second moment. Define $\hat\mu(n)= (1/n) \sum_{n' =1}^n X_{n'}$. For $\delta > 0$ define \newline $D^\delta(s) = \sum_{n=1}^s  \indic \{ |\hat\mu(n) - \mu| \geq \delta \}$. Then we have that for all $\delta$:
\eqs{
\frac{\EE[D^\delta(s)]} {\log(s)} \to_{s \to \infty} 0.
}
\end{lemma}
\bp
	We define $v^2 = \EE[( X_1 - \mu)^2 ]$ the variance. Using the fact that $\{X_n\}_{n \geq 1}$ are independent, we have that $\EE[(\hat\mu(n) - \mu)^2 ] = v^2/n$. Applying Chebychev's inequality we have that:
	\eqs{
	\PP[ |\hat\mu(n) - \mu| \geq \delta  ] \leq \frac{\EE[(\hat\mu(n) - \mu)^2 ]}{ \delta^2} = \frac{v^2}{n\delta^2}.
	}
	Therefore, we recognize the harmonic series: \eqs{
	\EE[D^\delta(s)] = \sum_{n=1}^s \PP[ |\hat\mu(n) - \mu| \geq \delta  ]  \leq \frac{v^2}{\delta^2} \sum_{n=1}^s \frac{1}{n} \leq \frac{v^2 (\log(s) + 1)}{\delta^2},
	}
	so that $\sup_s \EE[D^\delta(s)]/\log(s) < \infty$.
	
	Applying the law of large numbers, we have that $\hat\mu(n) \to_{n \to \infty} \mu$ a.s., so that $|\hat\mu(n) - \mu|$ occurs only finitely many times a.s. Hence $\sup_s D^\delta(s) < \infty$ a.s and $D^\delta(s)/\log(s) \to 0$ a.s.
	
	We have proven that $\sup_s \EE[D^\delta(s)]/\log(s) < \infty$ and $D^\delta(s)/\log(s) \to 0$ a.s. so applying Lebesgue's dominated convergence theorem we get the announced result:
\eqs{
\frac{\EE[D^\delta(s)]} {\log(s)} \to_{s \to \infty} 0,
}
which concludes the proof.	
\ep

\end{document}